\documentclass[twoside]{article}

\usepackage[T1]{fontenc}
\usepackage{xspace}
\usepackage{soul}
\usepackage{amsfonts}
\usepackage{url}
\usepackage{graphicx}
\usepackage{amsmath}
\usepackage{amsthm}
\usepackage{setspace}
\usepackage{natbib}
\usepackage{booktabs}
\usepackage[switch]{lineno}
\usepackage{hyperref}
\usepackage{cleveref}
\usepackage{multirow}
\usepackage{enumitem}
\usepackage{bbding}
\usepackage{comment}
\usepackage{subcaption}
\usepackage[noend]{algpseudocode}
\usepackage[T1]{fontenc}
\usepackage{flushend}
\usepackage[normalem]{ulem}
\usepackage{subfloat}
\usepackage{circledsteps}
\usepackage{adjustbox}
\usepackage{geometry}
\usepackage{float}
\usepackage{caption}
\usepackage{bm}
\usepackage{adjustbox}
\usepackage{makecell}
\usepackage{mathrsfs}
\usepackage{url}
\usepackage[linesnumbered,ruled,vlined]{algorithm2e}
\usepackage{color, xcolor}
\useunder{\uline}{\ul}{}

\newcommand{\method}{\textit{PENGUIN}\xspace}
\newcommand{\tian}[1]{\textcolor{blue}{#1}}

\usepackage[accepted]{aistats2026}
\begin{document}

%

%

\runningtitle{Enhancing Transformer with Periodic-Nested Group Attention for LTSF}

\runningauthor{Tian Sun, Yuqi Chen, Weiwei Sun}

\twocolumn[

\aistatstitle{PENGUIN: Enhancing Transformer with Periodic-Nested Group Attention for Long-term Time Series Forecasting}

\aistatsauthor{Tian Sun$^{1,2}$ \And Yuqi Chen$^{1,2,3}$ \And Weiwei Sun$^{1,2}$}

\aistatsaddress{$^1$ School of Computer Science, Fudan University}
\vspace{-8mm}
\aistatsaddress{$^2$ Shanghai Key Laboratory of Data Science, Fudan University}
\vspace{-8mm}
\aistatsaddress{$^3$ Xiaohongshu Inc.}
]

\newcommand{\hiddenfootnote}[1]{%
  \begingroup
  \stepcounter{footnote}%
  \renewcommand\thefootnote{}%
  \footnotemark%
  \footnotetext{\noindent #1}%
  \addtocounter{footnote}{-1}%
  \endgroup
}

\begin{abstract}

Despite advances in the Transformer architecture, their effectiveness for long-term time series forecasting (LTSF) remains controversial. In this paper, we investigate the potential of integrating explicit periodicity modeling into the self-attention mechanism to enhance the performance of Transformer-based architectures for LTSF. Specifically, we propose \method, a simple yet effective periodic-nested group attention mechanism. Our approach introduces a periodic-aware relative attention bias to directly capture periodic structures and a grouped multi-query attention mechanism to handle multiple coexisting periodicities (e.g., daily and weekly cycles) within time series data. Extensive experiments across diverse benchmarks demonstrate that \method consistently outperforms both MLP-based and Transformer-based models. Code is available at \tian{\url{https://github.com/ysygMhdxw/AISTATS2026_PENGUIN}}.\hiddenfootnote{Tian Sun and Yuqi Chen contributed equally to the research. Work was conducted during Yuqi Chen's affiliation with Fudan University. Correspondence to Weiwei Sun. contact: wwsun@fudan.edu.cn}

\end{abstract}

\section{Introduction}

Long-term time series forecasting (LTSF) is a pivotal task in various domains, including finance~\citep{chen2019investment,zhang2017stock}, traffic~\citep{zhang2018deeptravel,zheng2020gman,chen2023rntrajrec,sun2025learning}, and healthcare~\citep{chen2024eegformer,yi2024learning}, where accurate predictions of future values are essential for decision-making~\citep{zhou2022fedformer,zhang2023crossformer,wang2024timemixer2,zhang2024up2me,li2022learning,zhang2025enhancing}. Transformers, with their ability to capture long-range dependencies~\citep{vaswani2017attention}, have shown promising results in sequence-based tasks~\citep{touvron2023llama,bai2023qwen,brown2020language,raffel2020exploring}. However, their effectiveness in LTSF remains contentious~\citep{zeng2023transformers}. Recent studies suggest that simple linear models can outperform Transformer-based approaches~\citep{xu2024cyclenet,lin2024sparsetsf,xu2023fits}, raising questions about whether Transformers are effective for long-term time series forecasting.



In this paper, we rethink the design principle of self-attention and demonstrate that, when enhanced with \textbf{Pe}riodic-\textbf{N}ested \textbf{G}ro\textbf{u}p Attent\textbf{i}o\textbf{n} mechanism (short for \method), self-attention can be highly effective for LTSF. Specifically, designed to explicitly model temporal dependencies and intrinsic periodic patterns, \method incorporates two core components: a periodic-nested relative attention bias that directly captures periodic structures, and a grouped multi-query attention mechanism~\cite{ainslie2023gqa} that effectively handles multiple coexisting periodicities. The motivations behind this design can be summarized in the following two aspects.


Firstly, time series data often exhibit periodic patterns~\citep{xu2024cyclenet, lin2024sparsetsf, qin2024muse, wu2022timesnet, elfeky2005periodicity}, which conventional attention mechanisms struggle to capture effectively over long horizons~\citep{xu2024cyclenet,kim2024self,zeng2023transformers}. Existing methods, such as Autoformer~\citep{wu2021autoformer} and FEDformer~\citep{zhou2022fedformer}, attempt to model these patterns implicitly through frequency decomposition. However, recent work~\citep{xu2024cyclenet} suggests that explicitly modeling periodicity yields better results. Motivated by this, we pioneer integrating explicit periodicity modeling within the attention mechanism to enhance the LTSF performance, while retaining the strong modeling capability of Transformers.



Secondly, capturing multiple periodicities is essential for accurate time series forecasting, as real-world data often exhibits complex, overlapping temporal cycles~\citep{xu2024cyclenet,wang2024timemixer}. Existing methods, such as TimesNet~\citep{wu2022timesnet}, address this by decomposing time series into multiple temporal sequences, each corresponding to a different period length. Such an approach can be computationally inefficient. In contrast, we propose a group attention mechanism that partitions the multi-head attention into several groups, each dedicated to modeling a specific period length. To further improve efficiency, each group employs multi-query attention~\citep{shazeer2019fast}. This design facilitates more effective and specialized learning of diverse periodic patterns. In summary, the contributions of the paper are as follows:
\begin{itemize}
    \item  We pioneer the exploration of explicitly modeling periodicity within the attention structure to enhance the performance of the LTSF task.
    \item Technically, we introduce periodic-nested group attention, named \method, an effective and efficient approach to learn multiple periodicity patterns for time series data.
    \item Extensive experiments on benchmark datasets demonstrate that \method achieves state-of-the-art performance.
\end{itemize}

\section{Related Work}

\subsection{Transformers for Long-term Time Series Forecasting}


Applying Transformer architectures to time series forecasting has emerged as a prominent research topic within the time series community in recent years~\cite{zhou2021informer, liu2022pyraformer, liu2022non}. Concurrently, PatchTST~\citep{nie2022time} has established itself as the benchmark Transformer model by patching the time series input in a channel-independent manner~\citep{kim2024self}, a.k.a. CI strategy~\citep{shao2024exploring}. Following this strategy, PathFormer~\citep{chen2024pathformer} introduces a multi-scale Transformer architecture with adaptive pathways, partitioning time series data into multiple temporal resolutions. CATS~\citep{kim2024self} replaces self-attention with a cross-attention-only Transformer approach, highlighting the effectiveness of cross-attention for LTSF. Another strategy, known as the channel-dependent (CD) strategy, focuses on modeling the relationships among variates~\citep{liu2023itransformer}. iTransformer~\citep{liu2023itransformer} emerges as a straightforward yet effective implementation of the CD strategy, making no modification to the vanilla Transformer architecture~\citep{vaswani2017attention}.
Very recently, DGCformer~\citep{liu2024dgcformer} proposed relatively balanced channel strategies called channel-hard clustering (CHC), and DUET~\citep{qiu2024duet} adopted a channel-soft clustering (CSC) strategy and developed a fully adaptive sparsity module to dynamically build groups for each channel. 

In this paper, we intend to answer a fundamental question: Can periodic temporal patterns be explicitly modeled within the attention mechanism to enhance the performance for LTSF tasks? To this end, we observe that existing Transformers lack the ability to explicitly model multiple periodicities. Moreover, relative attention bias (RAB), successful in the field of natural language processing~\citep{raffel2020exploring,press2021train} and large language models~\citep{sturua2024jina}, has been under-exploited in time series forecasting. To overcome these limitations, we propose a simple yet effective periodic-nested group attention mechanism for long-term time series forecasting.

\subsection{Linear Models for Long-term Time Series Forecasting}

While Transformers have been extensively studied for LSTF, their effectiveness remains debatable. Recent works have demonstrated that simple linear models can outperform Transformer-based approaches~\citep{xu2024cyclenet,ni2024mixture}
DLinear~\citep{zeng2023transformers} proposes a one-layer linear model and demonstrates the effectiveness of simple architectures in LTSF. 
Both FITS~\citep{xu2023fits} and SparseTSF~\citep{lin2024sparsetsf} adopt lightweight designs, with FITS employing complex-valued neural networks and SparseTSF utilizing the sparse forecasting technique. 
Furthermore, CycleNet~\citep{xu2024cyclenet} explicitly models periodic patterns to capture cyclical dependencies in time series data. FreqMoE~\citep{liu2025freqmoe} introduces frequency decomposition to handle multiple periodicities implicitly.
These lightweight models, despite the risk of oversimplifying complex temporal patterns, achieve high forecasting accuracy, raising questions about the effectiveness of Transformers for LSTF. In this paper, we address this by enhancing the Transformer with Periodic-Nested Group Attention, achieving state-of-the-art performance.

\subsection{Language Model v.s. Time Series}

Transformers have demonstrated remarkable success in Large Language Models (LLMs)~\citep{yang2019xlnet}. Recently, several works have explored leveraging large language models for time series forecasting~\citep{liu2024unitime,ekambaram2024ttms,liu2024autotimes,cao2024tempo}. UniTime~\citep{liu2024unitime} proposes a language-empowered unified model for cross-domain time series forecasting. TEMPO~\citep{cao2024tempo} introduces prompt-based generative pre-trained Transformers for time series. These methods focus on leveraging LLM representations for TSF. However, unlike natural language, time series data exhibits a distinct periodic nature~\citep{xu2024cyclenet,fan2025pdg2seq,wu2021autoformer,wen2022transformers}. To bridge the gap, we extend the ALiBi bias~\citep{press2021train} by incorporating periodicity, enabling the model to capture repetitive seasonal trends\footnote{Although originally proposed to address input length extrapolation in natural language processing~\citep{press2021train}, ALiBi stands out for its simplicity, robustness, and strong performance even without extrapolation (i.e., the sequence length within the train dataset is equivalent to that in the test dataset), outperforming sinusoidal embedding~\citep{vaswani2017attention}. These advantages motivate our choice to extend ALiBi for time series data.}, making it fundamentally different from the LLM-based approaches.

\section{Preliminaries}

Long-term time series forecasting (LTSF) involves predicting future values of a time series based on historical observations. Formally, a multi-variate time series $X=[x_{t-L+1},\cdots, x_{t}] \in \mathbb{R}^{L \times C}$ represents a sequence of observations over $L$ timesteps, each with $C$ variates (a.k.a channels). The objective is to predict the future sequence $\overline{X} =[x_{t+1}, \cdots, x_{t+H} ] \in \mathbb{R}^{H \times C} $, where $H$ denotes the forecasting horizon.

\section{Proposed Methodology}


\begin{figure*}[htbp]
    \centering
    \includegraphics[width=0.85\textwidth]{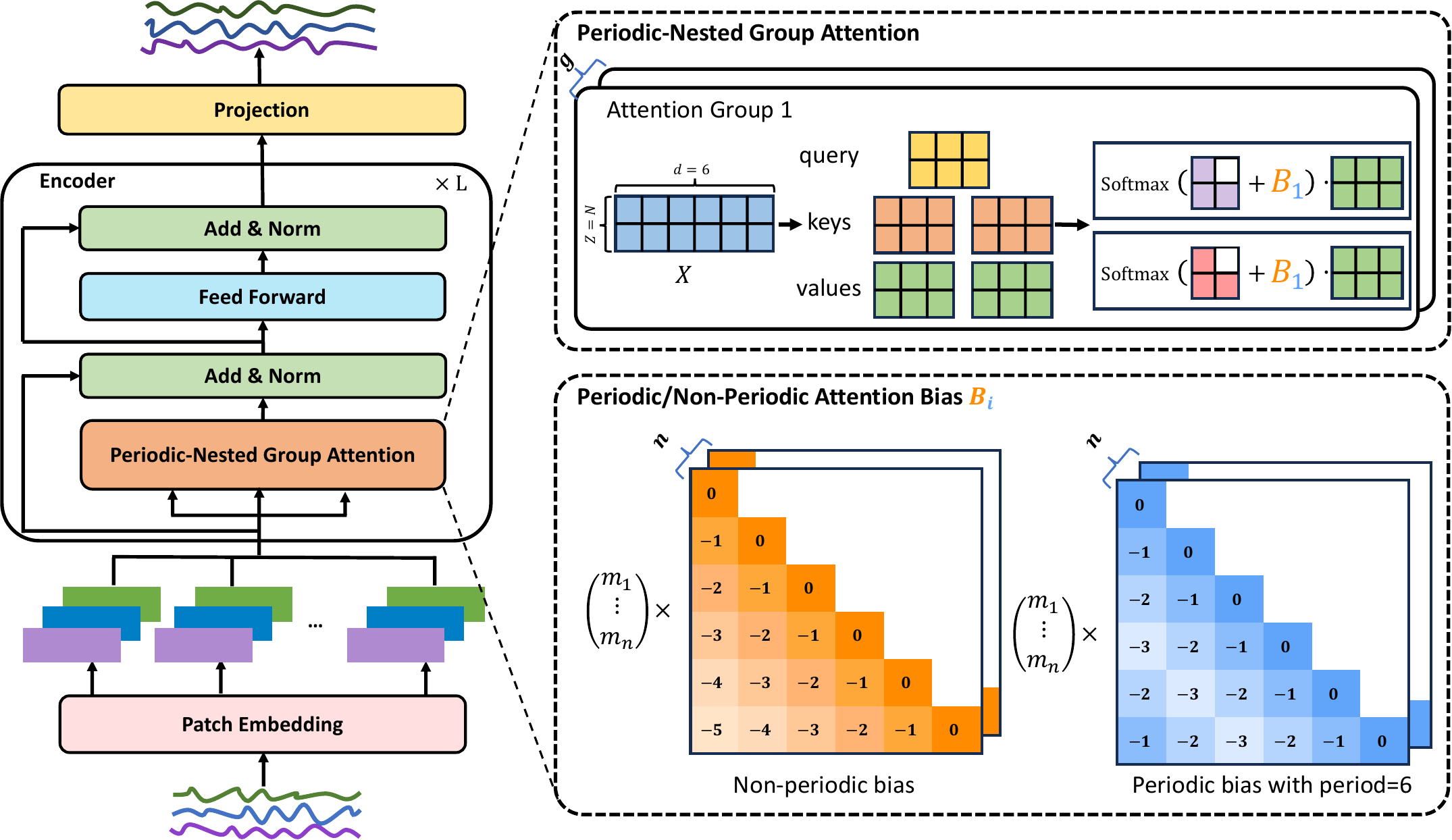}
    \caption{Illustration of \method, a novel Transformer model with periodic-nested group attention tailored for LTSF. We first transform time series into patch embeddings in a channel-independent manner. Next, the encoder module comprises the proposed periodic-nested group attention and a feed-forward network. The right part of the figure illustrates the periodic-nested group attention with periodic/non-periodic attention bias.
    }
    \label{fig:model}
\end{figure*}

We introduce \method, a novel periodic-nested group attention mechanism for accurate long-term time series forecasting. As illustrated in Figure~\ref{fig:model}, our model first converts time series into channel-independent patch representations via patch embedding. The encoder then leverages our enhanced self-attention with periodic-nested group attention to effectively capture periodic and temporal dependencies, followed by a linear projection to produce predictions. The model is optimized using mean squared error loss.

In the rest of this section, we will detail the proposed attention architecture, while leaving the details about input processing to Appendix ~\ref{sec:input}.



\subsection{Channel Independent Encoder}

The encoder processes the time series in a channel-independent manner. For the $c$-th channel, the input is first normalized via Revertible Instance Normalization~\citep{nie2022time} and projected to patch embeddings, yielding $x_e^{(c)} \in \mathbb{R}^{N \times d}$, where $N$ is the number of patches. The periodic information $\mathcal{P}_S$ (detailed in the next sub-section) is incorporated as input and shared uniformly across all channels. Suppose we have $E$ encoder layers. The overall equations for $l$-th layer are summarized as:
  \begin{align}
    x_e^{(c,l,1)} &= x_e^{(c, l-1)} + \text{Norm}\bigl(\text{PENGUIN}(x_e^{(c, l-1)};\mathcal{P}_S)\bigr) ~, \nonumber \\
    x_e^{(c,l,2)} &= x_e^{(c, l, 1)} + \text{Norm}\bigl(\text{FeedForward}(x_e^{(c, l, 1)})\bigr) ~,
  \end{align}
where $x_e^{(c, l)}=x_e^{(c,l,2)}$ and $x_e^{(c, 0)}=x_e^{(c)}$. Norm denotes the root mean square layer normalization~\citep{zhang2019root}, i.e.,
\begin{equation}
    \text{Norm}(x) = \frac{x}{\sqrt{\frac{1}{d} \sum_{i=1}^{d} x_i^2 + \epsilon}} \cdot \gamma ~,
\end{equation}
where $d$ is the dimensionality of $x$, and $\gamma \in \mathbb{R}^d$ are learnable weights. FeedForward is an MLP module, i.e.,
\begin{equation}
    \text{FeedForward}(X) = \text{ReLU}(XW_f^1 + b_f^1)W_f^2 + b_f^2 ~,
\end{equation}
where $W_f^1 \in \mathbb{R}^{d \times d_\text{ff}}$,$W_f^2 \in \mathbb{R}^{d_\text{ff} \times d}$,$ b_f^1 \in \mathbb{R}^{d_\text{ff}}$ and $ b_f^2 \in \mathbb{R}^{d}$ are learnable parameters, and $d_{\text{ff}}$ is the intermediate dimensionality of the feed-forward network. PENGUIN is the proposed periodic-nested group attention.

\subsection{Periodic-Nested Group Attention}

\method is an attention mechanism for capturing multiple periodicities within time series data. Specifically, given the input representation $X \in \mathbb{R}^{N \times d}$, we calculate self-attention in a group-query attention manner~\citep{ainslie2023gqa}. Assuming we have $h$ attention heads and $g$ attention groups. Additionally, we assume that $g$ is a factor of $h$. Therefore, the self-attention module applies linear projections to compute the query, key, and value matrices $Q \in \mathbb{R}^{N \times h \times d_h}$, $K \in \mathbb{R}^{N \times g \times d_h}$, and $V \in \mathbb{R}^{N \times g \times d_h}$, where $d_h = d / h$ denotes the dimensionality per head. Specifically, $Q = XW_Q$, $K = XW_K$, $V = XW_V$, where $W_Q \in \mathbb{R}^{d \times d}$, $W_K, W_V \in \mathbb{R}^{d \times (g \times d_h)}$ are trainable weight matrices. For each group, keys and values are shared, i.e., for each query within the $r$-th group, the group-specific key and value are denoted as $k^r \in \mathbb{R}^{d_h}$ and $v^r \in \mathbb{R}^{d_h}$~\citep{shazeer2019fast}.

Generally speaking, we denote $n=h/g$ for the number of heads per group. Each attention group calculates a multi-query attention with a group specified relative attention bias $\mathcal{B}_r \in \mathbb{R}^{n \times N \times N}$, $\mathcal{B}_r^{(k)}$ stands for the attention bias for the $k$ attention head within the group. Here, the attention bias is an additive term applied to the attention scores before softmax to encode structural priors such as temporal proximity and periodic alignment. Finally, for the $r$-th attention group with $n$ keys and values, the attention mechanism is formulated as:
\begin{align}
\text{head}_\ell &= \text{Softmax}\left(\frac{Q^\ell (k^r)^\top}{\sqrt{d_h}} + \mathcal{B}_{r}^{(k)} \right)v^r ~, \nonumber \\
\text{PENGUIN}_r(X) &= \text{Output}([\text{head}_{\tau+1}\| ...\| \text{head}_{\tau+n}]) ~,
\label{eq:overall}
\end{align}
where $\tau = (r-1)*n$, $r \in \{1, ..., g\}$, $\ell$ denotes the head index, $k = \ell \text{ Mod } n+1$, $[\cdot\|\cdot]$ indicates the concatenation operation, and the $\text{Output}(\cdot)$ denotes the linear projection. The final output of the attention module is obtained by concatenating the outputs from all $g$ groups, i.e., $\text{PENGUIN}(X) = [\text{PENGUIN}_1(X) \| \cdots \| \text{PENGUIN}_g(X)]$.


\begin{figure}[t]
    \centering
    \includegraphics[width=\linewidth]{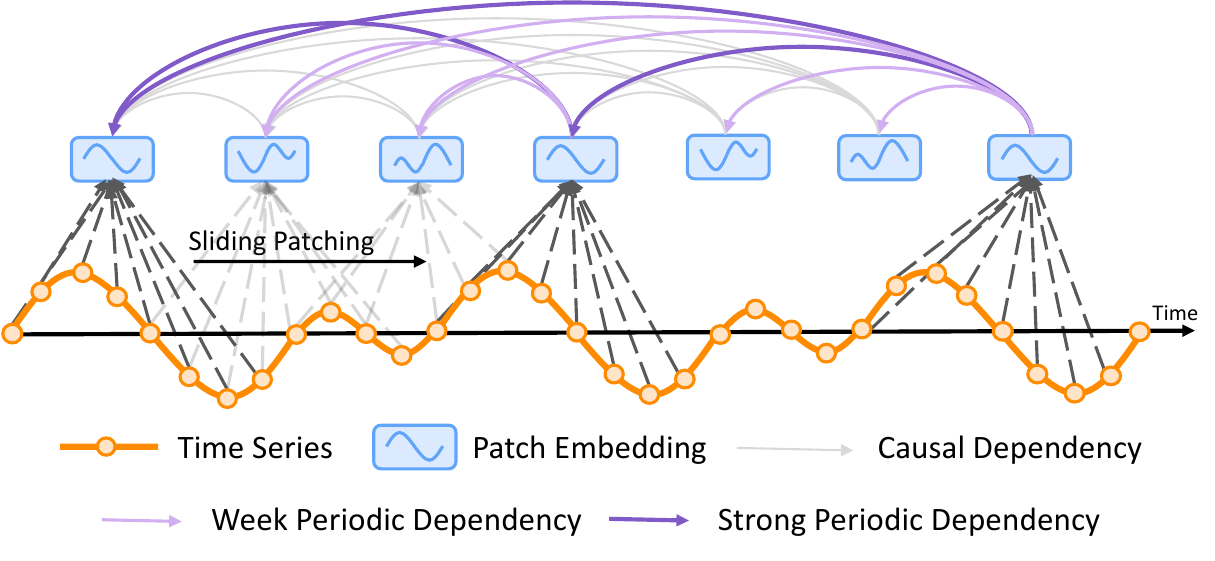}
    \caption{Illustration of periodic attention modeling with an overlapping patching technique. Given a time series with a period of 12, the patching length $P$ and stride $S$ are set to 8 and 4, respectively. Every three patch embeddings share the same sub-region of the original time series, leading to a period after patching of $\mathcal{P}_{S}=[3]$.}
    \label{fig:patch}
\end{figure}

\begin{table*}[t]
\begin{center}
\caption{Summary of datasets for long-term time series forecasting.}
\resizebox{\linewidth}{!}{
\begin{tabular}{c|c|c|c|c|c|c|c}
\toprule
Dataset & ETTh1 \& ETTh2 & ETTm1 \& ETTm2 & Electricity & Exchange & Weather & Solar & Traffic  \\
\midrule
Channels & 7 & 7 & 321 &8 &21&137&862\\
\midrule
Sampling Frequency & 1 hour & 15 mins & 1 hour &  1 Day  & 10 mins & 10 mins & 1 hour \\
\midrule
Natural Periodicity &  Daily & Daily & Daily \& Weekly & - & Daily & Daily & Daily \& Weekly \\
\midrule
Period Lengths $\mathcal{P}$ & $\{24\}$ & $\{96\}$ & $\{24,168\}$ & $\emptyset$ & $\{144\}$ & $\{144\}$ & $\{24,168\}$\\
\bottomrule
\end{tabular}
}
\label{tab:datasets}
\end{center}
\end{table*}

\subsubsection{Attention Bias for the Non-Periodic Case}

Time series data may contain instances where periodic information is absent, or the dataset itself lacks inherent periodicity. In such cases, prior methods such as CycleNet~\citep{xu2024cyclenet} often struggle to make accurate predictions and tend to degrade into simple linear predictors. Moreover, when time series lack inherent periodicity, frequency decomposition methods, such as Autoformer~\citep{wu2021autoformer} and FEDformer~\citep{zhou2022fedformer}, often produce highly inconsistent spectra across different windowed segments, undermining the stability and effectiveness of frequency-domain modeling. On the contrary, for the non-periodic case, we can define an effective linear bias dependent on the relative positional distance\footnote{For periodic data, such attention bias can still be beneficial. Therefore, we can also leave one "expert" or one attention group that does not rely on periodic information.}, i.e., 
\begin{equation}
\mathcal{B}_{ij}^{(k)}= -m_k \cdot |i-j| ~,
\label{eq:non-p}
\end{equation}
where $i,j \in [1,N]$, $m=[m_1, ..., m_n]$ is a head-specific slope fixed before training, with $m_k=2^{-\frac{8}{k}}$ and $k \in [1, n]$
One key benefit of the attention bias is that it encourages the model to focus more on local context while still maintaining access to long-range information. 

\subsubsection{Attention Bias for the Periodic Case}

Time series data exhibits periodic characteristics, distinguishing it from natural language data. To bridge the gap, we adopt a periodic-nested attention bias to capture periodic attributes from input embeddings adeptly. In this study, we assume that the periodic information or cycle lengths are known a priori for the dataset. Previous work~\citep{xu2024cyclenet} highlights that such periodic information could be obtained through autocorrelation functions (ACF)~\citep{cryer1986time}. Formally speaking, let the dataset be associated with multiple periods denoted as $\mathcal{P}=\{\mathcal{P}_1, ..., \mathcal{P}_p\}$, where $p$ denotes the number of periods. For instance, the Traffic dataset is collected hourly and exhibits both daily and weekly periodicity, corresponding to $\mathcal{P}=\{24, 168\}$.

Since the time series is transformed into a sequence of tokens through patching to reduce computational costs, the periodicity of the patch embeddings becomes dependent on the stride $S$. As illustrated in Figure \ref{fig:patch}, the periods after applying sliding patches are defined as $\mathcal{P}_{S}=\{\frac{\mathcal{P}_1}{S}, ..., \frac{\mathcal{P}_p}{S}\}$. Throughout this paper, we assume $S$ to be a factor of each element in $\mathcal{P}$.





Additionally, each attention group is specified for one particular period length. Let the period length after patching for the $r$-th attention group be defined as $\mathcal{P}_S^{(r)}$. Therefore, we can define a periodic-nested attention bias as:
\begin{align}
\mathcal{B}^{(k)}_{ij} &= -m_k \cdot \widehat{\mathcal{B}}^{(k)}_{ij} ~, \nonumber \\
\widehat{\mathcal{B}}^{(k)}_{ij} &= 
\begin{cases}
u, & \text{if } u < \frac{\mathcal{P}_S^{(r)}}{2}, \\
\mathcal{P}_S^{(r)} - u, & \text{if } u \geq \frac{\mathcal{P}_S^{(r)}}{2},
\end{cases}  \nonumber \\
\text{where } u &= |i - j| \text{ Mod } \mathcal{P}_S^{(r)} ~.
\label{eq:bias}
\end{align}
The modulo operation ensures that the bias reflects periodic variations aligned with the corresponding period value. Additionally, the bias slope vector, i.e., $m$, is shared across group, and we define $m=[m_1, ..., m_n]$, with $m_k=2^{-\frac{8}{k}}$ and $k \in [1, n]$.

\subsection{Output Projection and Loss Function}

We apply a linear projection to obtain predictions $\widetilde{X} = [\widetilde{x}_{t+1}, \dots, \widetilde{x}_{t+H}] \in \mathbb{R}^{H \times C}$. Following recent LSTF work, we use Mean Squared Error (MSE) loss. Let $\overline{X} = [x_{t+1}, \dots, x_{t+H}] \in \mathbb{R}^{H \times C}$ be the ground truth, the loss function is calculated as:
\begin{equation}
    \mathcal{L}_{\text{MSE}} = \frac{1}{H} \sum_{i=1}^H {\| \widetilde{x}_{t+i}-\overline{x}_{t+i} \|}_2^2.
\end{equation}

\section{Experiments}

\subsection{Experimental Setup}

\subsubsection{Datasets} 

To evaluate the effectiveness of \method, we conducted experiments on nine widely used datasets in the field of long-term time series forecasting. These datasets include ETT (comprising ETTh1, ETTh2, ETTm1, and ETTm2), Electricity, Exchange, Weather, Solar, and Traffic~\citep{wu2021autoformer, zhou2021informer, xu2024cyclenet}. A detailed overview of these datasets is provided in Table~\ref{tab:datasets}.


\begin{table*}[t]
\begin{center}
\caption{Multivariate long-term forecasting results. The look-back length $L$ is fixed
as $336$ and the results are averaged over prediction horizons of $H \in \{96, 192, 336, 720\}$. ETT indicates the averaged results of ETTh1, ETTh2, ETTm1, and ETTm2. The best results are in \textbf{bold} and the second best are underlined.}
\resizebox{0.95\textwidth}{!}{
\begin{tabular}{ccc|cc|cc|cc|cc|cc}
\toprule
Dataset & \multicolumn{2}{c|}{ETT} & \multicolumn{2}{c|}{Electricity}  & \multicolumn{2}{c|}{Exchange} & \multicolumn{2}{c|}{Weather} & \multicolumn{2}{c|}{Solar} & \multicolumn{2}{c}{Traffic}  \\
\cmidrule(r){2-3} \cmidrule(r){4-5} \cmidrule(r){6-7} \cmidrule(r){8-9} \cmidrule(r){10-11} \cmidrule(r){12-13} 
Metric & MSE & MAE &  MSE & MAE &  MSE & MAE &  MSE & MAE &  MSE & MAE &  MSE & MAE   \\
\midrule
Autoformer&0.451&0.461&0.229&0.340&0.732&0.634&0.348&0.391&0.913&0.703&0.679&0.422\\
FEDformer&0.408&0.439&0.251&0.355&0.769&0.643&0.333&0.382&0.324&0.404&0.622&0.386\\
DLinear&0.376&0.404&0.166&0.263&0.376&0.444&0.245&0.299&0.253&0.315&0.434&0.295\\
TimesNet&0.396&0.415&0.249&0.338&0.416&0.443&0.267&0.298&0.236&0.279&0.634&0.354\\
MoLE&0.380&0.404&0.166&0.264&0.557&0.512&0.244&0.298&0.253&0.314&0.434&0.295\\
MTLinear&0.377&0.404&0.166&0.263&\textbf{0.285}&\textbf{0.380}&0.246&0.300&0.252&0.314&0.434&0.295\\
PatchTST&0.356&0.390&0.179&0.278&0.448&0.443&0.234&0.270&\underline{0.203}&0.265&0.405&0.282\\
iTransformer&0.371&0.400&\underline{0.163}&\underline{0.258}&0.459&0.457&0.236&0.272&0.230&0.286&\underline{0.390}&\underline{0.275}\\
CATS&0.357&0.392&0.173&0.259&0.425&0.436&\textbf{0.227}&\textbf{0.266}&0.206&\textbf{0.248}&0.411&0.276 \\
SparseTSF&0.356&\underline{0.384}&0.174&0.266&0.397&0.427&0.250&0.282&0.266&0.281&0.436&0.290\\
CycleNet&\underline{0.351}&\underline{0.384}&\textbf{0.160}&\textbf{0.253}&0.400&0.422&0.246&0.279&0.223&0.277&0.423&0.288\\
\midrule
\method&\textbf{0.344}&\textbf{0.379}&0.165&0.261&\underline{0.340}&\underline{0.403}&\underline{0.228}&\underline{0.267}&\textbf{0.200}&\underline{0.253}&\textbf{0.388}&\textbf{0.262} \\

\bottomrule
\end{tabular}
}
\label{tab:main2}
\end{center}
\end{table*}

\begin{table*}[htbp]
\centering
\caption{Experiment results of different choices of periodic length. The look-back length $L$ is fixed as $336$ and the prediction horizons of $H \in \{96, 192, 336, 720\}$. Throughout this experiment, we set the number of attention heads to $12$ to ensure that the number of groups $g$ is a factor of $h$.}
\resizebox{0.98\linewidth}{!}{
\begin{tabular}{c|c|cc|cc|cc|cc|cc|cc|cc|cc}
\toprule
\multicolumn{2}{c}{Case} & \multicolumn{2}{c|}{No Bias} & \multicolumn{2}{c|}{Non-Periodic} & \multicolumn{6}{c|}{Periodic} & \multicolumn{6}{c}{Both} \\
\cmidrule(r){3-4} \cmidrule(r){5-6} \cmidrule(r){7-12} \cmidrule(r){13-18}

\multicolumn{2}{c}{$\mathcal{P}$} & \multicolumn{2}{c|}{$\emptyset$} & 
\multicolumn{2}{c|}{$\emptyset$} & \multicolumn{2}{c|}{$\{24\}$} & 
\multicolumn{2}{c|}{$\{168\}$} & \multicolumn{2}{c|}{$\{24,168\}$} &  \multicolumn{2}{c|}{$\{24\}$} & \multicolumn{2}{c|}{$\{168\}$} &
\multicolumn{2}{c}{$\{24,168\}$} \\

\cmidrule(r){3-4} \cmidrule(r){5-6} \cmidrule(r){7-8} \cmidrule(r){9-10} \cmidrule(r){11-12} \cmidrule(r){13-14} 
\cmidrule(r){15-16} \cmidrule(r){17-18}  
\multicolumn{2}{c}{Metric} & MSE & MAE & MSE & MAE & MSE & MAE & MSE & MAE & MSE & MAE & MSE & MAE & MSE & MAE & MSE & MAE  \\
\midrule
\multirow{5}{*}{\rotatebox{90}{Traffic}} 
&96&0.382&0.266&0.364&0.252&0.361&0.250&0.360&0.246&0.358&0.244&0.356&0.244&0.356&0.243&0.357&0.247\\
&192&0.401&0.270&0.385&0.256&0.378&0.254&0.380&0.259&0.377&0.255&0.378&0.256&0.378&0.254&0.376&0.253\\
&336&0.411&0.280&0.401&0.274&0.396&0.269&0.399&0.273&0.395&0.266&0.390&0.265&0.393&0.270&0.390&0.262\\
&720&0.437&0.290&0.434&0.286&0.433&0.289&0.429&0.289&0.433&0.288&0.428&0.285&0.429&0.284&0.428&0.286\\
\cmidrule(l){2-18}
& avg. &0.407&0.277&0.396&0.267&0.392&0.265&0.392&0.266&0.390&0.263&\underline{0.388}&\underline{0.262}&0.389&0.263&\textbf{0.387}&\textbf{0.262}\\
\midrule
\multirow{5}{*}{\rotatebox{90}{Electricity}} 
&96&0.143&0.246&0.144&0.247&0.138&0.236&0.136&0.235&0.137&0.236&0.139&0.237&0.146&0.249&0.138&0.247\\
&192&0.159&0.260&0.151&0.245&0.149&0.243&0.149&0.244&0.147&0.243&0.150&0.245&0.149&0.244&0.149&0.245\\
&336&0.176&0.276&0.175&0.274&0.170&0.267&0.171&0.269&0.169&0.266&0.173&0.261&0.172&0.269&0.171&0.268\\
&720&0.208&0.299&0.209&0.300&0.208&0.299&0.208&0.298&0.207&0.298&0.209&0.300&0.210&0.302&0.206&0.295\\
\cmidrule(l){2-18}
& avg. &0.171&0.270&0.169&0.266&0.166&0.261&\underline{0.166}&0.261&\textbf{0.165}&\textbf{0.260}&0.167&\underline{0.260}&0.169&0.266&0.166&0.263\\
\bottomrule

\end{tabular}}
\label{tab:biasmore}

\end{table*}

\subsubsection{Baselines} 

We selected the current state-of-the-art and representative works in this field as baselines. These include linear-based or MLP-based models (e.g., DLinear~\citep{zeng2023transformers}, SparseTSF~\citep{lin2024sparsetsf}, CycleNet~\citep{xu2024cyclenet}, MoLE~\citep{ni2024mixture}, MTLinear~\citep{nochumsohn2025multi}), Transformer-based models (e.g, Autoformer~\citep{wu2021autoformer}, FEDformer~\citep{zhou2022fedformer}, PatchTST~\citep{nie2022time}, iTransformer~\citep{liu2023itransformer}, CATS~\citep{kim2024self}), and CNN-based model (e.g., TimesNet~\citep{wu2022timesnet}).

\subsection{Main Results}

To demonstrate the effectiveness of \method in capturing long-range periodic patterns, we evaluate its performance across various forecast horizons while keeping the input sequence length fixed at $336$. Table~\ref{tab:main2} presents a comparison of \method against other models on multivariate long-term series forecasting (LTSF) tasks. \footnote{To ensure fair comparison, we reproduce baseline results using the code and scripts from TimesLib. We set the input length to 336 to align with our experimental setting and avoid the "drop-last" issue that discards samples in the final batch during evaluation.} A detailed result can be found in Table~\ref{tab:main}. Moreover, we provide a comprehensive evaluation using the input sequence length of $96$ as shown in Table~\ref{tab:main96}. From the results, we can draw the following four conclusions:

\noindent \textit{(i) State-of-the-art performance}: \method achieves state-of-the-art overall results. Specifically, compared to the state-of-the-art MLP-based model CycleNet~\citep{xu2024cyclenet}, \method achieves an overall improvement of \textbf{5.3\%} (0.317 $\rightarrow$ \textbf{0.300}) in terms of MSE. Moreover, when compared to the leading Transformer-based model CATS~\citep{kim2024self}, \method delivers an overall MSE improvement of \textbf{6.0\%} (0.319 $\rightarrow$ \textbf{0.300}).

\noindent \textit{(ii) Superior improvement over existing decomposition approaches}: \method significantly outperforms Autoformer~\citep{wu2021autoformer} and FEDformer~\citep{zhou2022fedformer} by over \textbf{45\%} in MSE, underscoring the importance of explicitly modeling periodic information to boost performance.

\noindent \textit{(iii) Consistent improvement over Transformer-based approaches}: \method outperforms PatchTST~\citep{nie2022time} and CATS~\citep{kim2024self} across almost all datasets, except for the Electricity dataset, where CATS achieves better performance. We hypothesize that this is due to the Electricity dataset being more affected by non-stationarity~\citep{shao2024exploring,liu2024timebridge}, which \method is not specifically designed to handle. In conclusion, these results highlight the robustness of \method and its superior capacity compared to existing Transformer-based approaches.

\noindent \textit{(iv) Effective periodic modeling technique}: Both \method and CycleNet~\citep{xu2024cyclenet} aim to capture periodic patterns in time series data. The superior performance of \method over CycleNet demonstrates its effectiveness and offers novel insights into modeling periodic patterns for time series data.

\begin{table}[t]
\centering
\caption{Experiment results of MHA compared to GQA. The look-back length $L$ is fixed as $336$ and the results are averaged over prediction horizons of $H \in \{96, 192, 336, 720\}$. Time denotes the average duration of a single training iteration (seconds/iter).}
\resizebox{\linewidth}{!}{
\begin{tabular}{c|ccc|ccc}
\toprule
Dataset & \multicolumn{3}{c|}{Traffic} & \multicolumn{3}{c}{Electricity} \\
\cmidrule(r){2-4} \cmidrule(r){5-7} 
Metric & Time & MSE & MAE & Time & MSE & MAE  \\
\midrule
MHA &0.166&0.389&0.263&0.239&0.167&0.264\\
GQA &\textbf{0.145}&\textbf{0.387}&\textbf{0.262}&\textbf{0.224}&\textbf{0.165}&\textbf{0.260}\\
Improve & +12.7\% & +0.51\% & +0.38\% & +6.28\% & +1.20\% & +1.51\% \\
\bottomrule

\end{tabular}}
\label{tab:gqamore}

\end{table}

\subsection{Ablation Study}

\subsubsection{Ablation Study of Attention Bias}

In this paper, we propose a simple yet effective attention mechanism, termed \method, specifically designed for LTSF. To validate its effectiveness, we conduct a comprehensive ablation study on the Traffic and Electricity datasets, as shown in Table~\ref{tab:biasmore}. Notably, both datasets exhibit multiple periodic patterns~\citep{xu2024cyclenet}, including daily and weekly cycles. We conduct four groups of experiments: (i) No Bias, where we adopt multi-query attention with a single attention group~\citep{shazeer2019fast}; (ii) Non-Periodic, where we also adopt multi-query attention with a single attention group, but with a non-periodic attention bias; (iii) Periodic, where we adopt a group-query attention with various predefined periodic information; (iv) Both, where we combine both types of biases by assigning one attention group to non-periodic bias and the remaining groups to periodic biases. From this table, we can draw the following four conclusions:

\noindent (i) The attention mechanism without any bias yields the poorest performance, underscoring the importance of incorporating attention bias in LTSF. For instance, a non-periodic bias improves the performance by 2.77\% (0.407 $\rightarrow$ 0.396) on the Traffic dataset.

\noindent (ii) Results from the periodic groups consistently outperform their non-periodic counterparts, highlighting the critical role of periodic information in long-term forecasting. For instance, a multiple-periodic bias further improves the performance by 2.32\% (0.396 $\rightarrow$ 0.387) on the Traffic dataset.

\noindent (iii) When multiple periodicities are present in the dataset, leveraging these patterns leads to an overall 0.5\% performance improvement on both datasets.

\noindent (iv) Periodic attention bias captures recurring patterns, while non-periodic bias focuses on preserving temporal dependencies. Our empirical study shows that combining both types of attention bias yields improved performance on the Traffic dataset.

\subsubsection{Ablative Study of Attention Structure}

To improve \method's efficiency, we replace standard multi-head attention (MHA) with group-query attention (GQA), where queries remain head-specific but keys and values are shared within each attention group. As reported in Table~\ref{tab:gqamore}, GQA lowers computational overhead, cutting time cost by 13\% on the Traffic dataset, and consistently yields performance gains across multiple benchmarks. These results highlight GQA as an effective trade-off between efficiency and accuracy in long-term time series forecasting.

\begin{table}[t]
\centering
\caption{Experiment results of incorporating \method with RCF technique. The look-back length $L$ is fixed as $336$ and the prediction horizons of $H \in \{96, 192, 336, 720\}$.}
\resizebox{\linewidth}{!}{
\begin{tabular}{c|cccccc}
\toprule
Dataset & \multicolumn{2}{c}{Traffic} & \multicolumn{2}{c}{Solar} & \multicolumn{2}{c}{Electricity}  \\
\cmidrule(r){2-3} \cmidrule(r){4-5} \cmidrule(r){6-7} 
Metric & MSE & MAE & MSE & MAE & MSE & MAE \\
\midrule
CycleNet & 0.423 & 0.288 & 0.223 & 0.277 & \textbf{0.160} & \textbf{0.265} \\
\midrule
\method  &\textbf{0.388}&\textbf{0.262}&0.200&\textbf{0.253}&0.165&0.261\\
+ RCF  &0.401&0.268&\textbf{0.198}&0.254&0.163&0.258\\
Improve & -3.35\% & -2.29\%&+1.00\%&-0.39\%&+1.21\%&+1.15\%\\
\bottomrule
\end{tabular}}

\label{tab:rcfmore}

\end{table}

\subsubsection{Ablation of Periodic Patterns Modeling}

Both \method and the Residual Cycle Forecasting (RCF) technique~\citep{xu2024cyclenet} are designed to capture periodic patterns in time series data. While \method focuses on modeling complex and implicit periodic dependencies, RCF decomposes the time series into a learnable recurrent cycle component and a residual component, where the cycle component explicitly captures periodic patterns by learning a fixed-length repeating template. To assess their complementary effectiveness, we integrate RCF into \method and report the results in Table~\ref{tab:rcfmore}, comparing the original \method, the combined \method+RCF~\citep{xu2024cyclenet}, and CycleNet. The results show only marginal performance gains from incorporating RCF with \method. This suggests that the Transformer-based \method is sufficiently capable of modeling periodic dependencies without additional reliance on the RCF components.

\begin{table}[t]
\centering
\caption{The effectiveness of \method for Decoder architecture. The look-back length $L$ is fixed as $336$ and the results are averaged over prediction horizons of $H \in \{96, 192, 336, 720\}$.}
\resizebox{\linewidth}{!}{
\begin{tabular}{c|cccccc}
\toprule
Dataset & \multicolumn{2}{c}{ETTh1} & \multicolumn{2}{c}{ETTm1} & \multicolumn{2}{c}{Weather}  \\
\cmidrule(r){2-3} \cmidrule(r){4-5} \cmidrule(r){6-7} 
Metric & MSE & MAE & MSE & MAE & MSE & MAE \\
\midrule
Decoder  &0.478&0.465&0.448&0.439&0.228&0.261\\
+ \method  &\textbf{0.430}&\textbf{0.440}&\textbf{0.361}&\textbf{0.392}&\textbf{0.226}&\textbf{0.258}\\
Improve &+10.00\%&+5.43\%&+19.36\%&+10.60\%&+1.23\%&+1.04\%\\
\midrule
EncDec &0.495&0.473&0.520&0.469&0.236&0.267\\
+ \method &\textbf{0.419}&\textbf{0.436}&\textbf{0.354}&\textbf{0.388}&\textbf{0.231}&\textbf{0.265} \\
Improve  &+15.42\%&+7.88\%&+31.96\%&+17.33\%&+2.04\%&+0.68\%\\
\bottomrule
\end{tabular}}

\label{tab:decoder}

\end{table}

\begin{figure}[t]
    \centering
    \includegraphics[width=\linewidth]{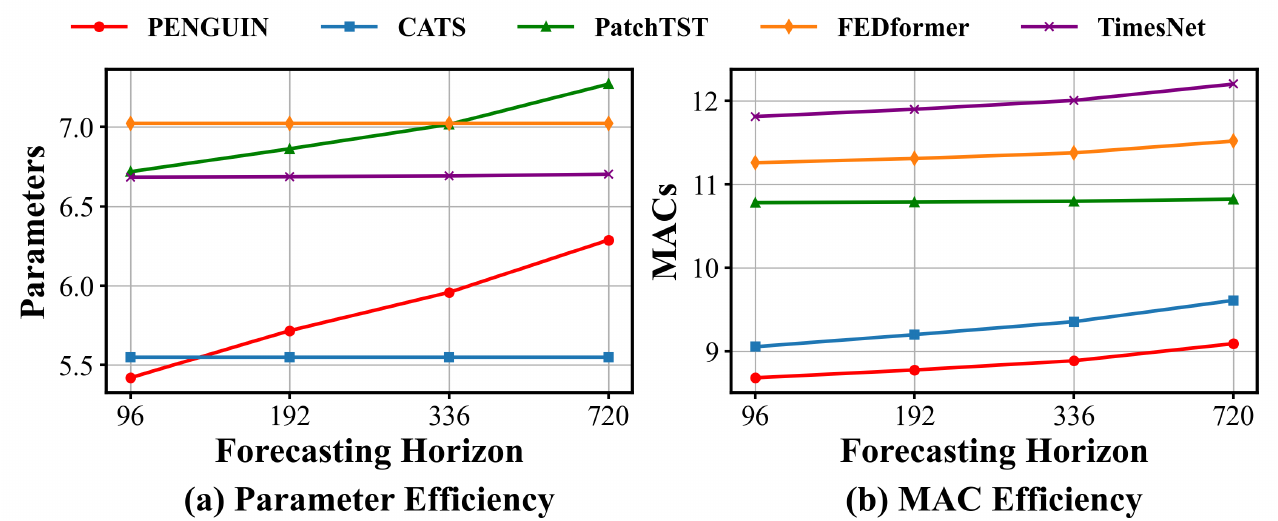}
    \caption{The efficiency analysis of five models (i.e., PatchTST~\citep{nie2022time}, CATS~\citep{kim2024self}, TimesNet~\citep{wu2022timesnet}, FEDformer~\citep{zhou2022fedformer}, and \method) on ETTm1 dataset. The experiments are conducted with a look-back length $L$ of $336$. Both the number of parameters and MACs are plotted on a logarithmic scale.}
    \label{fig:efficiency}
\end{figure}

\begin{figure}[ht]
    \centering
    \includegraphics[width=\linewidth]{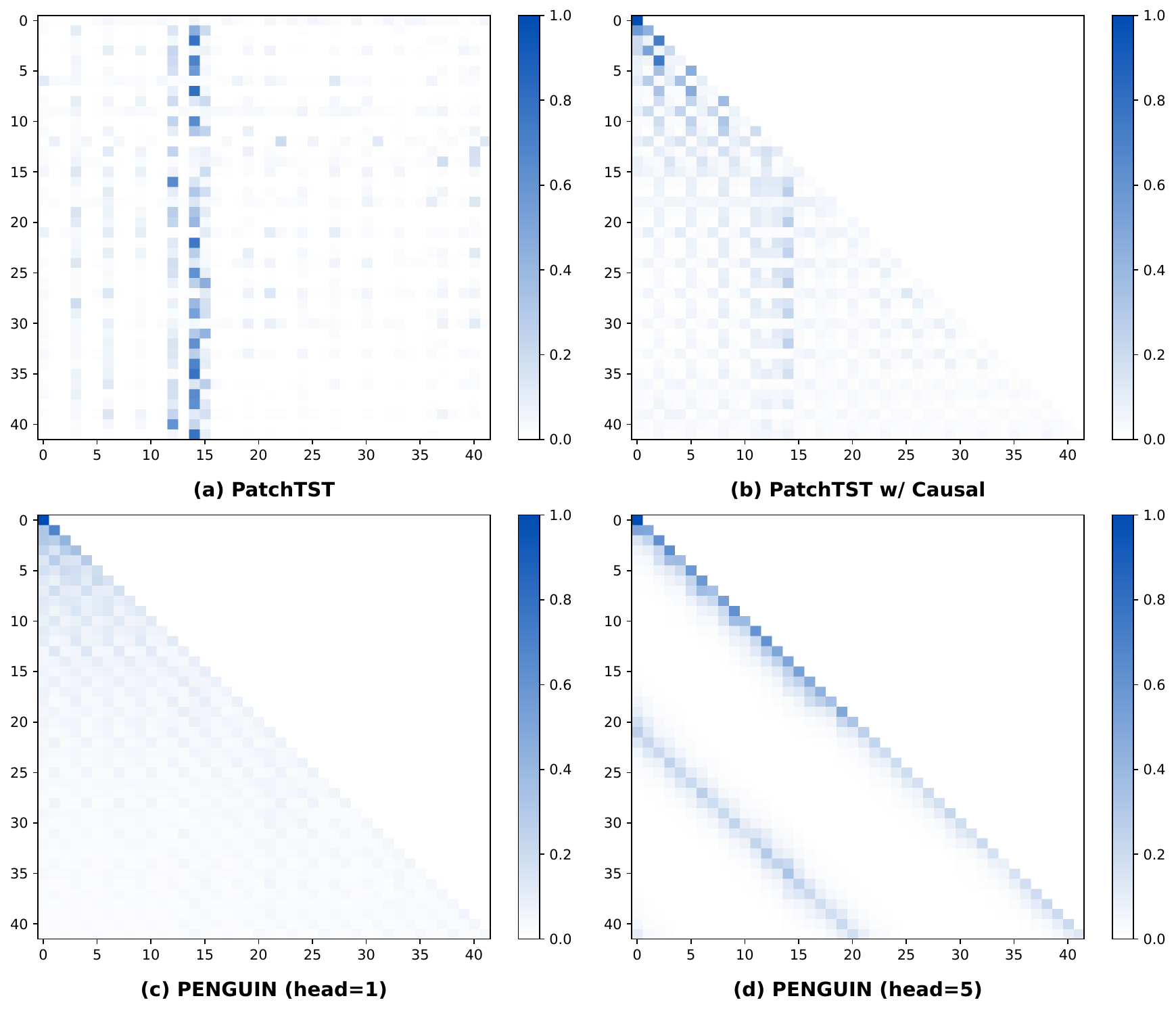}
    \caption{Visualization of \method and PatchTST on the Traffic dataset, using $\mathcal{P} = \{24, 168\}$ with a patch length $P = 16$ and stride $S = 8$. For PatchTST, we train two variants with different masking strategies (i.e., full attention and attention with a causal mask). For \method, we visualize two attention heads, each belonging to a different attention group.}
    \label{fig:heatmap}
\end{figure}

\subsection{Extendability Study}

The Transformer consists of encoder and decoder modules~\citep{vaswani2017attention}. We extend \method to the decoder, with a key difference in the non-periodic attention bias: $\mathcal{B}_{ij}^{(k)}= m_k \cdot |(i+N)-j|$, replacing Eq.~\eqref{eq:non-p}. Here, $N$ is the length of keys and values in the encoder architecture. \footnote {In cross-attention forecasting, keys and values come from the input series of length $N$, so $M$ decoding tokens align with positions $N+1,…,N+M$~\citep{kim2024self}.} A similar modification is applied to Eq. \eqref{eq:bias} for the periodic case. The experimental results, presented in Table~\ref{tab:decoder}, reveal that \method delivers significant improvements for both the decoder structure and the encoder-decoder (EncDec) models. These findings suggest that \method has strong potential as a plugin for Transformer-based models in time series modeling.

\subsection{Efficiency Study}

To assess the efficiency of \method, we consider two key metrics: the number of parameters and Multiply-Accumulate Operations (MACs). The parameter count measures the total number of learnable weights, while MACs quantify the computational cost of inference. The results, evaluated on the ETTm1 dataset with a fixed batch size of $32$ across all models, are presented in Figure~\ref{fig:efficiency}. For comparison, we include four representative baselines designed to model temporal dependencies in time series data. Notably, \method achieves the highest prediction accuracy, outperforming the second-best baseline, CATS, by 1.5\%, while simultaneously requiring the lowest computational cost among all evaluated approaches. In particular, \method demonstrates a substantial advantage over TimesNet in terms of MAC efficiency, underscoring the inefficiency of decomposing time series into multiple temporal sequences.


\begin{table*}[t]
\centering
\caption{Experiment results of missing or incorrect periodic information. The look-back length $L$ is fixed as $336$ and the prediction horizons of $H \in \{96, 192, 336, 720\}$.}
\resizebox{0.95\linewidth}{!}{
\begin{tabular}{c|c|cc|cc|cc|cc|cc|cc|cc|cc}
\toprule
\multicolumn{2}{c}{Case} & \multicolumn{4}{c}{Existing Approaches} & \multicolumn{12}{c}{\method} \\
\cmidrule(r){3-6} \cmidrule(r){7-18}
\multicolumn{2}{c}{Model / $\mathcal{P}$} & \multicolumn{2}{c|}{SparseTSF} & \multicolumn{2}{c|}{CATS} & \multicolumn{2}{c|}{$\emptyset$} &  \multicolumn{2}{c|}{$\{24\}$} & \multicolumn{2}{c|}{$\{72\}$} & \multicolumn{2}{c|}{$\{96\}$} & \multicolumn{2}{c|}{$\{144\}$} & \multicolumn{2}{c}{$\{168\}$}  \\
\cmidrule(r){3-4} \cmidrule(r){5-6} \cmidrule(r){7-8} \cmidrule(r){9-10}
\cmidrule(r){11-12} \cmidrule(r){13-14} \cmidrule(r){15-16} \cmidrule(r){17-18}
\multicolumn{2}{c}{Metric} & MSE & MAE & MSE & MAE & MSE & MAE & MSE & MAE & MSE & MAE & MSE & MAE & MSE & MAE & MSE & MAE \\
\midrule
\multirow{5}{*}{\rotatebox{90}{Traffic}} 
&96&0.415&0.281&0.382&0.260&0.364&0.252&0.361&0.250&0.366&0.254&0.363&0.250&0.361&0.251&0.360&0.246 \\
&192&0.426&0.283&0.401&0.268&0.385&0.256&0.378&0.254&0.383&0.258&0.383&0.262&0.382&0.261&0.380&0.259 \\
&336&0.438&0.289&0.414&0.275&0.401&0.274&0.396&0.269&0.400&0.274&0.402&0.274&0.401&0.276&0.399&0.273 \\
&720&0.465&0.306&0.447&0.300&0.434&0.286&0.433&0.289&0.433&0.292&0.433&0.290&0.432&0.291&0.429&0.289 \\
\cmidrule(l){2-18}
& avg. &0.436&0.290&0.411&0.276&0.396&0.267&\textbf{0.392}&\textbf{0.266}&0.396&0.270&0.395&0.269&0.394&0.270&0.392&0.267 \\
\midrule
\multirow{5}{*}{\rotatebox{90}{ETTm1}} 
&96&0.306&0.347&0.297&0.352&0.290&0.341&0.288&0.341&0.290&0.341&0.290&0.341&0.290&0.344&0.290&0.342 \\
&192&0.339&0.366&0.323&0.369&0.330&0.370&0.328&0.365&0.329&0.367&0.319&0.361&0.329&0.368&0.328&0.366 \\
&336&0.374&0.386&0.357&0.389&0.361&0.390&0.359&0.389&0.363&0.391&0.355&0.387&0.362&0.391&0.360&0.390 \\
&720&0.428&0.416&0.413&0.421&0.414&0.420&0.411&0.423&0.415&0.427&0.409&0.419&0.414&0.421&0.413&0.420 \\
\cmidrule(l){2-18}
& avg. &0.362&0.379&0.348&0.383&0.349&0.380&0.347&0.380&0.349&0.382&\textbf{0.343}&\textbf{0.377}&0.349&0.381&0.348&0.380 \\
\bottomrule

\end{tabular}}
\label{tab:missingmore}

\end{table*}

\subsection{Visualization Study}

We compare the attention matrices of \method and PatchTST~\citep{nie2022time}, as shown in Figure~\ref{fig:heatmap}. In the figure, the $i$-th row denotes the attention distribution over all tokens conditioned on the $i$-th query token. As illustrated in Figures~\ref{fig:heatmap}(c)–(d), \method effectively captures periodic temporal correlations, with different attention heads or groups learning distinct periodic patterns. In contrast, PatchTST produces less interpretable attention maps, while its causal-attention variant reveals some structural patterns but still falls short of \method. We hypothesize that the causal attention mechanism is essential for preserving temporal causality and modeling the auto-regressive generative nature of time series data. Supporting this claim, our empirical study shows that removing the causal mask from \method leads to performance drops of 3.5\% and 4.8\% on ETTh1 and ETTh2, respectively. Together, these results underscore the importance of explicitly modeling temporal dependencies and highlight the effectiveness of \method in capturing both causal and periodic structures for accurate time series forecasting.

\subsection{Robustness Towards Missing and Incorrect Periodic Information}

In real-world time series forecasting tasks, it is common for periodic information to be either missing or incorrectly specified due to noisy data or inaccurate prior knowledge. To evaluate \method's robustness in such scenarios, we conducted experiments by intentionally introducing missing or incorrect periodic information into the datasets~\citep{xu2024cyclenet}. Specifically, we select various periodic lengths from [24,72,96,144,168]. The results, shown in Table~\ref{tab:missingmore}, compare \method with representative approaches, i.e., SparseTSF and CATS. We can draw the following two conclusions:

\noindent (i) \method maintains comparable, and sometimes state-of-the-art, performance against strong baselines even when periodic information is missing, highlighting its ability to capture temporal dependencies without relying on precise periodic cues.

\noindent (ii) When periodic information is incorrect, performance degrades noticeably as expected (e.g., 1.7\% on the ETTm1 dataset), since incorrect patterns hinder model predictions. However, even with incorrect periodicity, \method outperforms many existing models (e.g., SparseTSF), demonstrating its ability to extract useful temporal features despite periodic discrepancies.

\section{Conclusion}

Long-term time series forecasting (LSTF) is a fundamental task with broad real-world applications. In this paper, we revisit the effectiveness of self-attention mechanisms and demonstrate that incorporating a simple attention bias can significantly enhance their capability for LSTF. To this end, we propose a novel periodic-nested group attention mechanism, \method, which combines a periodic linear bias with a grouped query attention structure. This design enables the model to capture diverse periodic patterns while maintaining temporal causality. Extensive experiments on nine benchmark datasets show that \method consistently achieves state-of-the-art performance across a wide range of LSTF tasks.

\begin{singlespace}\bibliographystyle{chicago}
\phantomsection\addcontentsline{toc}{section}{\refname}\bibliography{references}
\end{singlespace}

\section*{Checklist}



\begin{enumerate}

  \item For all models and algorithms presented, check if you include:
  \begin{enumerate}
    \item A clear description of the mathematical setting, assumptions, algorithm, and/or model. [Yes]
    \item An analysis of the properties and complexity (time, space, sample size) of any algorithm. [Yes]
    \item (Optional) Anonymized source code, with specification of all dependencies, including external libraries. [Yes]
  \end{enumerate}

  \item For any theoretical claim, check if you include:
  \begin{enumerate}
    \item Statements of the full set of assumptions of all theoretical results. [Not Applicable]
    \item Complete proofs of all theoretical results. [Not Applicable]
    \item Clear explanations of any assumptions. [Not Applicable]     
  \end{enumerate}

  \item For all figures and tables that present empirical results, check if you include:
  \begin{enumerate}
    \item The code, data, and instructions needed to reproduce the main experimental results (either in the supplemental material or as a URL). [Yes]
    \item All the training details (e.g., data splits, hyperparameters, how they were chosen). [Yes]
    \item A clear definition of the specific measure or statistics and error bars (e.g., with respect to the random seed after running experiments multiple times). [Yes]
    \item A description of the computing infrastructure used. (e.g., type of GPUs, internal cluster, or cloud provider). [Yes]
  \end{enumerate}

  \item If you are using existing assets (e.g., code, data, models) or curating/releasing new assets, check if you include:
  \begin{enumerate}
    \item Citations of the creator If your work uses existing assets. [Yes]
    \item The license information of the assets, if applicable. [Yes]
    \item New assets either in the supplemental material or as a URL, if applicable. [Yes]
    \item Information about consent from data providers/curators. [Yes]
    \item Discussion of sensible content if applicable, e.g., personally identifiable information or offensive content. [Yes]
  \end{enumerate}

  \item If you used crowdsourcing or conducted research with human subjects, check if you include:
  \begin{enumerate}
    \item The full text of instructions given to participants and screenshots. [Not Applicable]
    \item Descriptions of potential participant risks, with links to Institutional Review Board (IRB) approvals if applicable. [Not Applicable]
    \item The estimated hourly wage paid to participants and the total amount spent on participant compensation. [Not Applicable]
  \end{enumerate}

\end{enumerate}

\clearpage
\appendix
\thispagestyle{empty}

\onecolumn
\aistatstitle{Enhancing Transformer with Periodic-Nested Group Attention for Long-term Time Series Forecasting - Supplementary Materials}

\section{Details about Input Processing} \label{sec:input}

\subsection{Revertible Instance Normalization}

Time series data often encounter challenges related to distributional shifts in terms of statistical properties such as mean and variance over time~\citep{nie2022time,liu2023itransformer}.
To address the challenge, a standard normalization strategy, namely reversible instance normalization (Revin)~\citep{kim2021reversible}, demonstrates a great enhancement for LTSF. Mathematically, given the input $X \in \mathbb{R}^{L \times C}$, Revin normalizes the data by removing its mean and variance:
\begin{equation}
    x^{(c)}=\frac{x^{(c)}-\mu^{(c)}}{\sqrt{\sigma^{(c)}+\epsilon}}, c \in \{1, \ldots, C\} ~,
\end{equation}
where $\mu^{(c)}$ and $\sigma^{(c)}$ are the mean and standard deviation across the $c$-th channel input sequence, and $\epsilon$ is a small constant added for numerical stability. 
After the model processes the data, the original mean and variance are restored, mitigating the impact of non-stationary components. This approach improves the model's robustness and accuracy.

\subsection{Input Patching}

Long-term time series possess extensive look-back windows, necessitating models that can adeptly capture both local and long-term information. This enhancement through patching has been substantiated in numerous studies~\citep{nie2022time}. This approach involves dividing each univariate input time series $x^{(c)} \in \mathbb{R}^{L}$, where $c$ represents the $c$-th channel, into patches that serve as either overlapping or non-overlapping input tokens. Formally, given an input sequence of length $L$, the input time series sequence is divided into non-overlapping subsequences of length $P$ with a stride of $S$, resulting in $N=\lfloor\frac{(L-P)}{S}\rfloor+2$ patches as $x_p^{(c)} \in  \mathbb{R}^{N \times P}$. Subsequently, we embed each patch into a higher-dimensional space, $\tilde{x}_p^{(c)}= x_p^{(c)} W_p  +b_p$, where $W_p \in \mathbb{R}^{P\times d }$ and $b_p \in \mathbb{R}^{d}$ denote the learnable embedding matrix and bias vector, $d$ represents the size of hidden dimension. To incorporate the sequential order of the patches, a position embedding $PE$ is added to enhance the model's ability to understand the absolute position of each patch within the sequence. This process is mathematically represented as $x_e^{(c)} = \tilde{x}_p^{(c)} + PE$, where $x_e^{(c)} \in \mathbb{R}^{N \times d}$ represents the final input to the Transformer encoder and $PE$ is the position embedding vector that encodes the position information.

\section{Details about applying \method for Decoder Structure}

In the vanilla Transformer, the decoder network contains a self-attention sub-module and a cross-attention sub-module. To incorporate \method for the decoder structure, we extend the proposed attention mechanism for the cross-attention sub-module. Furthermore, following the cross-attention architecture of CATS~\citep{kim2024self}, the input to the decoder model consists of learnable queries. Since the time series is transformed to a patch embedding corresponding to the patch stride of $S$, the number of learning queries is $M=\frac{H}{S}$ and the input for the $c$-th channel is $x_d^{(c)} \in \mathbb{R}^{M \times d}$. After processing through the decoder model, the output $\tilde{x}_d^{(c)} \in \mathbb{R}^{M \times d}$ is transformed to the output via linear transformation, i.e., $\widetilde{x}_d^{(c)}=\tilde{x}_d^{(c)} W_d^T + b_d$, where $W_d \in \mathbb{R}^{d \times S}$ and $b_d \in \mathbb{R}^{S}$. Afterward, the transformed output $\widetilde{x}_d^{(c)} \in \mathbb{R}^{M \times S}$ is then concatenated to reconstruct the final sequence $\widetilde{x}^{(c)} \in \mathbb{R}^{H}$. 

\paragraph{Attention Bias for the Periodic Case for Cross Attention} Unlike the encoder, the number of tokens for queries and keys can be different. Specifically, given queries $Q \in \mathbb{R}^{M \times d}$ and keys $K \in \mathbb{R}^{N \times d}$, the periodic bias in Equation~\eqref{eq:bias} should be modified as follows:

\begin{align}
\mathcal{B}^{(k)}_{ij} &= -m_k \cdot \widehat{\mathcal{B}}^{(k)}_{ij} ~, \nonumber \\
\widehat{\mathcal{B}}^{(k)}_{ij} &= 
\begin{cases}
u, & \text{if } u < \frac{\mathcal{P}_S^{(r)}}{2}, \\
\mathcal{P}_S^{(r)} - u, & \text{if } u \geq \frac{\mathcal{P}_S^{(r)}}{2},
\end{cases}  \nonumber \\
\text{where } u &= |(i+N) - j| \text{ Mod } \mathcal{P}_S^{(r)} ~.
\label{eq:bias}
\end{align}
Since there are $N$ tokens corresponding to the keys (i.e., the past history), the first decoding token — that is, the first query — should be assigned an index of $N+1$, the second $N+2$, and so on. Consequently, the positional bias between the $i$-th query and the $j$-th key should be defined as $|(i+N)-j|$ instead of $|i-j|$, reflecting the fact that queries represent future tokens while keys represent past tokens.

Furthermore, the decoder’s objective is to predict future horizons, with each learnable query token responsible for forecasting $S$ future values. As a result, causal masking is no longer necessary in the cross-attention mechanism, since future tokens are allowed unrestricted access to information from past tokens.

\section{Details about Experimental Setting}

\subsection{Dataset Split and Hyper-parameter Settings}

Following the settings of \citet{nie2022time} and \citet{xu2024cyclenet}, we split the ETT datasets (ETTh1, ETTh2, ETTm1, ETTm2) into training, validation, and testing sets using a 6:2:2 ratio, while the other datasets are divided using a 7:1:2 ratio. For our model, we adopt fixed positional embeddings for the input data~\citep{vaswani2017attention}. To address GPU memory limitations during training, we adapt the batch size according to the dataset characteristics: larger batch sizes are used for datasets with fewer feature channels, while smaller batch sizes are applied to higher-dimensional datasets to ensure stable optimization~\citep{lin2024sparsetsf}. A comprehensive overview of the hyperparameter configurations and training setups is provided in Table~\ref{tab:setting}.

\begin{table*}[htbp]
\begin{center}
\resizebox{0.8\textwidth}{!}{
\begin{tabular}{c|c|c|c|c|c|c}
\toprule
Dataset & Batch Size & Hidden Dimension & Patch Length & Patch Stride & Epochs & Patience \\
\midrule
ETTh1 & 64 & 16&1&1&30&6\\ 
ETTh2 &32 &16&12&2&30&6\\
ETTm1 & 64&16&4&2&30&6\\
ETTm2 &32 &16&2&2&30&6\\
Electricity & 12&128&12&2&30&6\\
Weather & 8&128&4&1&30&6\\
Solar &4&512&16&1&30&10\\
Traffic &4&512&16&8&10&6\\
\bottomrule
\end{tabular}
}
\end{center}
\caption{Experimental setting of various time series forecasting datasets.}
\label{tab:setting}
\end{table*}

\subsection{Hyper-Parameter Search Space}

We observe that patch stride, patch length, and hidden dimensions influence the performance. Therefore, we conduct the following search space as shown in Table~\ref{tab:search}.

\begin{table}[htbp]
\centering
\caption{Hyperparameter search space of \method.}
\label{tab:hyperparam-search}
\resizebox{0.4\linewidth}{!}{
\begin{tabular}{cc}
\toprule
Hyperparameter       & Values \\
\midrule
Patch Stride         & [1, 2, 4, 8] \\
Patch Length         & [1, 2, 4, 8, 12, 16] \\
Hidden Dimensions    & [16, 64, 128, 512] \\
\bottomrule
\end{tabular}}
\label{tab:search}
\end{table}

To avoid the OOM issue during hyperparameter search, we set the batch size according to the GPU limitation when using the largest hyperparameter.

\subsection{Compared Baselines}

We selected the current state-of-the-art and representative works in this field as baselines. These include linear-based or MLP-based models, including:
\begin{itemize}
    \item DLinear: A combination of a decomposition scheme~\citep{wu2021autoformer,zhou2022fedformer} with linear layers.
    \item SparseTSF~\citep{lin2024sparsetsf}: A linear approach that simplifies the forecasting task by decoupling the periodicity and trend in time series data. 
    \item CycleNet~\citep{xu2024cyclenet}: The state-of-the-art MLP-based approach for LTSF, modeling time series through residual cycle forecasting (RCF) to capture inherent periodic patterns\footnote{For datasets without periodicity (e.g., Exchange), we remove the RCF technique from CycleNet, resulting in a model with a single linear layer and reversible instance normalization.}.
    \item MoLE~\citep{ni2024mixture}: A Mixture-of-Linear-Experts approach that employs channel-wise MoE with a linear-based model for TSF.
    \item MTLinear~\citep{nochumsohn2025multi}: A multi-task learning approach for TSF where similar variates are grouped together, and each group forms a separate task.
\end{itemize}

Additionally, we also compare the Transformer-based models, including:
\begin{itemize}
    \item Autoformer~\citep{wu2021autoformer}: A novel decomposition architecture incorporating an Auto-Correlation mechanism, which empowers Autoformer with progressive decomposition capabilities for modeling complex time series.
    \item FEDformer~\citep{zhou2022fedformer}: An efficient Transformer design that segments time series into patches and applies shared embeddings and weights across univariate series.
    \item PatchTST~\citep{nie2022time}: An approach that first divides time series into sub-level patches, followed by a Transformer network.
    \item iTransformer~\citep{liu2023itransformer} (short for iTrans.): An approach that applies the attention and feed-forward network on the inverted dimensions to capture multivariate correlations.
    \item CATS~\citep{kim2024self}: The state-of-the-art Transformer-based approach for LTSF that eliminates self-attention, leveraging cross-attention mechanisms instead.
\end{itemize}

Furthermore, we compare \method with a convolution-based approach, TimesNet~\citep{wu2022timesnet}, an approach that transforms time series data into 2D tensors and captures multi-periodic patterns through an inception block.

\subsection{Metrics and Environments} 

Consistent with prevalent methodologies~\citep{nie2022time,xu2024cyclenet,zeng2023transformers}, we elect the traditional Mean Squared Error (MSE) and Mean Absolute Error (MAE) as the core metrics for evaluation. 

\subsection{Environments} All experiments are conducted on a single NVIDIA RTX 3090 GPU with 24GB of memory.



\section{Additional Experimental Results}

\subsection{Main Results}

To demonstrate the effectiveness of \method in capturing long-range periodic patterns, we evaluate its performance across various forecast horizons while keeping the input sequence length fixed at $336$. Table~\ref{tab:main} presents a comparison of \method against other models on multivariate long-term series forecasting (LTSF) tasks. As reported in the table, \method achieves state-of-the-art results, attaining the highest top-1 count and the lowest average error across benchmarks. Moreover, table~\ref{tab:main96} summarizes the experimental results using the input sequence length fixed at $96$, a standard experimental setting for most research on long-term time series forecasting. Notably, the results of CycleNet~\cite{xu2024cyclenet}, iTransformer~\citep{liu2023itransformer}, and CATS~\citep{kim2024self} are directly copied from the original paper. The results of PatchTST~\citep{nie2022time} are copied from the iTransformer~\citep{liu2023itransformer} paper to avoid the drop-last issue of time series evaluation. The results show that PENGUIN achieves state-of-the-art performance as compared to recent strong baselines.

{\color{blue} 
\begin{table*}[htbp]
\centering
\caption{Multivariate long-term forecasting results using a fixed random seed across benchmarks and compared models. The look-back length $L$ is fixed as $336$ and the results are averaged over prediction horizons of $H \in \{96, 192, 336, 720\}$. The best results are in \textbf{bold} and the second best are \underline{underlined}.}
\adjustbox{max width=\textwidth, max totalheight=0.85\textheight}{%
\begin{tabular}{c|c|cc|cc|cc|cc|cc|cc|cc|cc|cc|cc}
\toprule
\multicolumn{2}{c|}{Model} & \multicolumn{2}{c|}{ETTh1} & \multicolumn{2}{c|}{ETTh2} & \multicolumn{2}{c|}{ETTm1} & \multicolumn{2}{c|}{ETTm2} & \multicolumn{2}{c|}{Electricity} & \multicolumn{2}{c|}{Exchange} & \multicolumn{2}{c|}{Weather} & \multicolumn{2}{c|}{Solar} & \multicolumn{2}{c|}{Traffic} & \multicolumn{2}{c}{\textbf{Avg.}} \\
\cmidrule(lr){3-4} \cmidrule(lr){5-6} \cmidrule(lr){7-8} \cmidrule(lr){9-10} \cmidrule(lr){11-12} \cmidrule(lr){13-14} \cmidrule(lr){15-16} \cmidrule(lr){17-18} \cmidrule(lr){19-20} \cmidrule(lr){21-22}
\multicolumn{2}{c|}{Metric} & MSE & MAE & MSE & MAE & MSE & MAE & MSE & MAE & MSE & MAE & MSE & MAE & MSE & MAE & MSE & MAE & MSE & MAE & MSE & MAE \\
\midrule
\multirow{4}{*}{\rotatebox{90}{\method}} & 96 & 0.375 & 0.397 & 0.283 & \textbf{0.343} & \textbf{0.290} & \textbf{0.341} & 0.168 & \underline{0.258} & 0.137 & 0.236 & 0.093 & 0.218 & \textbf{0.150} & \underline{0.202} & \textbf{0.180} & \underline{0.244} & \textbf{0.357} & \textbf{0.247} & \multirow{4}{*}{\textbf{0.300}} & \multirow{4}{*}{\textbf{0.330}} \\
 & 192 & 0.420 & 0.426 & \underline{0.351} & \textbf{0.387} & \textbf{0.319} & \textbf{0.361} & \underline{0.227} & \textbf{0.296} & \underline{0.147} & 0.243 & 0.190 & 0.316 & \underline{0.194} & \textbf{0.243} & \textbf{0.200} & \underline{0.256} & \textbf{0.376} & \textbf{0.253} &  &  \\
 & 336 & \underline{0.436} & 0.439 & 0.376 & \textbf{0.409} & \textbf{0.355} & 0.387 & 0.282 & \textbf{0.333} & 0.169 & 0.266 & \underline{0.313} & \textbf{0.408} & \underline{0.245} & \underline{0.284} & \textbf{0.205} & \textbf{0.254} & \textbf{0.390} & \textbf{0.262} &  &  \\
 & 720 & \textbf{0.448} & \textbf{0.461} & \textbf{0.403} & 0.438 & \textbf{0.409} & 0.419 & \textbf{0.374} & \textbf{0.390} & 0.207 & 0.298 & \underline{0.763} & \underline{0.668} & \underline{0.323} & \underline{0.339} & \underline{0.214} & \underline{0.259} & \underline{0.428} & \textbf{0.286} &  &  \\
\midrule
\multirow{4}{*}{\rotatebox{90}{CATS}} & 96 & 0.381 & 0.399 & 0.294 & 0.350 & 0.297 & 0.352 & 0.197 & 0.289 & \textbf{0.130} & \textbf{0.218} & 0.101 & 0.222 & \underline{0.151} & 0.203 & \underline{0.191} & \textbf{0.237} & 0.382 & \underline{0.260} & \multirow{4}{*}{0.319} & \multirow{4}{*}{\underline{0.339}} \\
 & 192 & 0.421 & 0.432 & 0.356 & 0.391 & \underline{0.323} & 0.369 & 0.252 & 0.325 & 0.147 & \textbf{0.236} & \textbf{0.182} & \textbf{0.305} & \textbf{0.193} & \underline{0.243} & 0.203 & \textbf{0.246} & 0.401 & \underline{0.268} &  &  \\
 & 336 & \textbf{0.432} & 0.444 & 0.387 & 0.416 & \underline{0.357} & 0.389 & 0.304 & 0.357 & \underline{0.165} & \textbf{0.253} & 0.342 & 0.427 & \textbf{0.244} & \textbf{0.282} & 0.214 & \underline{0.256} & 0.414 & \underline{0.275} &  &  \\
 & 720 & 0.495 & 0.486 & 0.415 & 0.443 & \underline{0.413} & 0.421 & 0.395 & 0.411 & 0.251 & 0.327 & 1.074 & 0.790 & \textbf{0.319} & \textbf{0.335} & 0.214 & \textbf{0.251} & 0.447 & 0.300 &  &  \\
\midrule
\multirow{4}{*}{\rotatebox{90}{CycleNet}} & 96 & 0.382 & 0.403 & \textbf{0.281} & \underline{0.345} & 0.301 & \underline{0.344} & 0.172 & 0.263 & \underline{0.130} & \underline{0.226} & 0.091 & \underline{0.211} & 0.174 & 0.225 & 0.203 & 0.287 & 0.399 & 0.276 & \multirow{4}{*}{\underline{0.317}} & \multirow{4}{*}{0.339} \\
 & 192 & 0.414 & 0.422 & \textbf{0.349} & 0.392 & 0.335 & \underline{0.364} & \textbf{0.226} & \underline{0.299} & \textbf{0.146} & \underline{0.241} & 0.192 & 0.310 & 0.216 & 0.259 & 0.224 & 0.297 & 0.413 & 0.282 &  &  \\
 & 336 & 0.438 & \textbf{0.437} & \underline{0.373} & 0.415 & 0.369 & \textbf{0.384} & \textbf{0.280} & 0.335 & \textbf{0.162} & \underline{0.257} & 0.385 & 0.449 & 0.263 & 0.293 & 0.227 & 0.261 & 0.426 & 0.288 &  &  \\
 & 720 & \underline{0.455} & \underline{0.468} & 0.428 & 0.453 & 0.424 & \textbf{0.414} & 0.381 & 0.395 & \underline{0.200} & \textbf{0.289} & 0.930 & 0.719 & 0.330 & 0.339 & 0.238 & 0.264 & 0.453 & 0.305 &  &  \\
\midrule
\multirow{4}{*}{\rotatebox{90}{SparseTSF}} & 96 & 0.425 & 0.424 & 0.290 & 0.345 & 0.306 & 0.347 & 0.174 & 0.264 & 0.149 & 0.243 & 0.101 & 0.226 & 0.177 & 0.227 & 0.234 & 0.264 & 0.415 & 0.281 & \multirow{4}{*}{0.327} & \multirow{4}{*}{0.343} \\
 & 192 & 0.438 & 0.431 & 0.352 & \underline{0.387} & 0.339 & 0.366 & 0.231 & 0.302 & 0.160 & 0.253 & 0.189 & 0.313 & 0.220 & 0.262 & 0.263 & 0.279 & 0.426 & 0.283 &  &  \\
 & 336 & 0.441 & \underline{0.438} & \textbf{0.372} & 0.413 & 0.374 & \underline{0.386} & \underline{0.280} & \underline{0.333} & 0.175 & 0.268 & 0.334 & 0.420 & 0.267 & 0.297 & 0.284 & 0.291 & 0.438 & 0.289 &  &  \\
 & 720 & 0.462 & 0.474 & \underline{0.404} & \textbf{0.434} & 0.428 & \underline{0.416} & \underline{0.375} & \underline{0.390} & 0.213 & 0.300 & 0.963 & 0.749 & 0.335 & 0.343 & 0.283 & 0.290 & 0.465 & 0.306 &  &  \\
\midrule
\multirow{4}{*}{\rotatebox{90}{iTrans.}} & 96 & 0.404 & 0.420 & 0.301 & 0.358 & 0.302 & 0.357 & 0.173 & 0.266 & 0.133 & 0.228 & 0.103 & 0.230 & 0.159 & 0.209 & 0.201 & 0.256 & \underline{0.365} & 0.261 & \multirow{4}{*}{0.329} & \multirow{4}{*}{0.350} \\
 & 192 & 0.451 & 0.450 & 0.376 & 0.405 & 0.350 & 0.385 & 0.247 & 0.314 & 0.156 & 0.251 & 0.212 & 0.334 & 0.203 & 0.251 & 0.223 & 0.278 & \underline{0.378} & 0.269 &  &  \\
 & 336 & 0.476 & 0.467 & 0.398 & 0.423 & 0.380 & 0.402 & 0.297 & 0.345 & 0.167 & 0.265 & 0.369 & 0.448 & 0.254 & 0.289 & 0.243 & 0.298 & \underline{0.392} & 0.277 &  &  \\
 & 720 & 0.538 & 0.523 & 0.416 & 0.444 & 0.446 & 0.442 & 0.379 & 0.398 & \textbf{0.195} & \underline{0.289} & 1.153 & 0.815 & 0.327 & 0.340 & 0.254 & 0.312 & \textbf{0.424} & \underline{0.293} &  &  \\
\midrule
\multirow{4}{*}{\rotatebox{90}{TimesNet}} & 96 & 0.384 & 0.402 & 0.393 & 0.418 & 0.342 & 0.387 & 0.198 & 0.280 & 0.209 & 0.312 & 0.107 & 0.234 & 0.179 & 0.230 & 0.223 & 0.268 & 0.607 & 0.344 & \multirow{4}{*}{0.376} & \multirow{4}{*}{0.375} \\
 & 192 & 0.436 & 0.429 & 0.416 & 0.441 & 0.374 & 0.407 & 0.297 & 0.340 & 0.242 & 0.331 & 0.226 & 0.344 & 0.241 & 0.283 & 0.242 & 0.294 & 0.625 & 0.349 &  &  \\
 & 336 & 0.491 & 0.469 & 0.403 & 0.439 & 0.417 & 0.428 & 0.327 & 0.365 & 0.257 & 0.343 & 0.367 & 0.448 & 0.297 & 0.322 & 0.236 & 0.281 & 0.641 & 0.356 &  &  \\
 & 720 & 0.521 & 0.500 & 0.460 & 0.475 & 0.469 & 0.460 & 0.408 & 0.403 & 0.288 & 0.367 & 0.964 & 0.746 & 0.350 & 0.356 & 0.242 & 0.273 & 0.663 & 0.367 &  &  \\
\midrule
\multirow{4}{*}{\rotatebox{90}{PatchTST}} & 96 & 0.380 & 0.406 & 0.288 & 0.346 & \underline{0.290} & 0.345 & 0.174 & 0.265 & 0.142 & 0.245 & 0.100 & 0.227 & 0.151 & \textbf{0.199} & 0.193 & 0.253 & 0.377 & 0.270 & \multirow{4}{*}{0.321} & \multirow{4}{*}{0.344} \\
 & 192 & 0.422 & 0.435 & 0.358 & 0.390 & 0.340 & 0.379 & 0.229 & 0.301 & 0.178 & 0.280 & 0.185 & 0.313 & 0.200 & 0.249 & \underline{0.200} & 0.264 & 0.394 & 0.275 &  &  \\
 & 336 & 0.473 & 0.464 & 0.381 & \underline{0.412} & 0.374 & 0.400 & 0.288 & 0.337 & 0.186 & 0.288 & 0.392 & 0.451 & 0.252 & 0.287 & \underline{0.209} & 0.274 & 0.407 & 0.283 &  &  \\
 & 720 & 0.483 & 0.489 & 0.405 & \underline{0.437} & 0.435 & 0.439 & 0.378 & 0.391 & 0.208 & 0.299 & 1.113 & 0.780 & 0.334 & 0.344 & \textbf{0.210} & 0.269 & 0.442 & 0.299 &  &  \\
\midrule
\multirow{4}{*}{\rotatebox{90}{MTLinear}} & 96 & \textbf{0.371} & \textbf{0.392} & 0.309 & 0.368 & 0.301 & 0.345 & 0.170 & 0.264 & 0.140 & 0.237 & \textbf{0.086} & \textbf{0.208} & 0.178 & 0.243 & 0.222 & 0.291 & 0.410 & 0.282 & \multirow{4}{*}{0.321} & \multirow{4}{*}{0.352} \\
 & 192 & \textbf{0.408} & \textbf{0.416} & 0.352 & 0.396 & 0.336 & 0.368 & 0.235 & 0.315 & 0.154 & 0.250 & \underline{0.182} & \underline{0.308} & 0.216 & 0.275 & 0.249 & 0.312 & 0.423 & 0.287 &  &  \\
 & 336 & 0.455 & 0.454 & 0.423 & 0.446 & 0.381 & 0.402 & 0.303 & 0.365 & 0.169 & 0.267 & \textbf{0.306} & \underline{0.416} & 0.265 & 0.319 & 0.269 & 0.326 & 0.436 & 0.296 &  &  \\
 & 720 & 0.500 & 0.510 & 0.634 & 0.565 & 0.447 & 0.443 & 0.407 & 0.424 & 0.203 & 0.300 & \textbf{0.565} & \textbf{0.586} & 0.325 & 0.363 & 0.271 & 0.328 & 0.466 & 0.315 &  &  \\
\midrule
\multirow{4}{*}{\rotatebox{90}{MoLE}} & 96 & \underline{0.371} & \underline{0.393} & 0.291 & 0.354 & 0.299 & 0.344 & \underline{0.167} & 0.262 & 0.140 & 0.238 & \underline{0.088} & 0.213 & 0.175 & 0.237 & 0.222 & 0.291 & 0.410 & 0.282 & \multirow{4}{*}{0.353} & \multirow{4}{*}{0.367} \\
 & 192 & \underline{0.409} & \underline{0.420} & 0.357 & 0.396 & 0.336 & 0.367 & 0.236 & 0.318 & 0.154 & 0.250 & 0.488 & 0.491 & 0.216 & 0.275 & 0.250 & 0.312 & 0.423 & 0.288 &  &  \\
 & 336 & 0.446 & 0.447 & 0.441 & 0.458 & 0.374 & 0.393 & 0.304 & 0.360 & 0.169 & 0.268 & 0.599 & 0.576 & 0.260 & 0.310 & 0.268 & 0.325 & 0.437 & 0.297 &  &  \\
 & 720 & 0.495 & 0.506 & 0.721 & 0.604 & 0.439 & 0.438 & 0.396 & 0.416 & 0.203 & 0.301 & 1.053 & 0.770 & 0.327 & 0.369 & 0.271 & 0.329 & 0.466 & 0.315 &  &  \\
\midrule
\multirow{4}{*}{\rotatebox{90}{DLinear}} & 96 & 0.374 & 0.398 & \underline{0.281} & 0.347 & 0.307 & 0.350 & \textbf{0.165} & \textbf{0.250} & 0.140 & 0.237 & 0.110 & 0.252 & 0.174 & 0.235 & 0.222 & 0.292 & 0.410 & 0.282 & \multirow{4}{*}{0.331} & \multirow{4}{*}{0.359} \\
 & 192 & 0.430 & 0.440 & 0.367 & 0.404 & 0.340 & 0.373 & 0.227 & 0.307 & 0.153 & 0.250 & 0.208 & 0.349 & 0.219 & 0.281 & 0.249 & 0.313 & 0.423 & 0.288 &  &  \\
 & 336 & 0.442 & 0.445 & 0.438 & 0.454 & 0.377 & 0.397 & 0.304 & 0.362 & 0.169 & 0.267 & 0.410 & 0.501 & 0.264 & 0.317 & 0.268 & 0.327 & 0.436 & 0.296 &  &  \\
 & 720 & 0.497 & 0.507 & 0.598 & 0.549 & 0.433 & 0.433 & 0.431 & 0.441 & 0.203 & 0.299 & 0.774 & 0.675 & 0.324 & 0.363 & 0.271 & 0.326 & 0.466 & 0.315 &  &  \\
\midrule
\multirow{4}{*}{\rotatebox{90}{FEDformer}} & 96 & 0.387 & 0.433 & 0.383 & 0.430 & 0.382 & 0.428 & 0.256 & 0.333 & 0.225 & 0.334 & 0.375 & 0.454 & 0.276 & 0.345 & 0.275 & 0.374 & 0.606 & 0.380 & \multirow{4}{*}{0.437} & \multirow{4}{*}{0.436} \\
 & 192 & 0.435 & 0.461 & 0.420 & 0.457 & 0.404 & 0.441 & 0.290 & 0.353 & 0.232 & 0.342 & 0.490 & 0.524 & 0.309 & 0.370 & 0.308 & 0.382 & 0.620 & 0.383 &  &  \\
 & 336 & 0.463 & 0.474 & 0.430 & 0.468 & 0.479 & 0.474 & 0.334 & 0.380 & 0.259 & 0.361 & 0.715 & 0.645 & 0.353 & 0.395 & 0.376 & 0.446 & 0.623 & 0.386 &  &  \\
 & 720 & 0.492 & 0.503 & 0.479 & 0.493 & 0.471 & 0.466 & 0.428 & 0.436 & 0.286 & 0.382 & 1.494 & 0.948 & 0.395 & 0.419 & 0.337 & 0.414 & 0.640 & 0.396 &  &  \\
\midrule
\multirow{4}{*}{\rotatebox{90}{Autoformer}} & 96 & 0.551 & 0.499 & 0.381 & 0.427 & 0.459 & 0.463 & 0.266 & 0.341 & 0.208 & 0.323 & 0.351 & 0.445 & 0.294 & 0.369 & 1.019 & 0.753 & 0.679 & 0.419 & \multirow{4}{*}{0.523} & \multirow{4}{*}{0.481} \\
 & 192 & 0.514 & 0.505 & 0.398 & 0.437 & 0.550 & 0.512 & 0.298 & 0.361 & 0.210 & 0.327 & 0.542 & 0.559 & 0.341 & 0.388 & 0.904 & 0.706 & 0.677 & 0.417 &  &  \\
 & 336 & 0.501 & 0.511 & 0.393 & 0.437 & 0.608 & 0.548 & 0.331 & 0.378 & 0.218 & 0.333 & 0.661 & 0.624 & 0.372 & 0.411 & 0.964 & 0.710 & 0.663 & 0.415 &  &  \\
 & 720 & 0.513 & 0.505 & 0.448 & 0.475 & 0.590 & 0.549 & 0.414 & 0.428 & 0.280 & 0.376 & 1.373 & 0.908 & 0.384 & 0.394 & 0.763 & 0.642 & 0.697 & 0.437 &  &  \\
\bottomrule
\end{tabular}}
\label{tab:main}
\end{table*}
}

\begin{table*}[htbp]
\begin{center}
\caption{Multivariate long-term forecasting results. The look-back length $L$ is fixed
as $96$ and the results are averaged over prediction horizons of $H \in \{96, 192, 336, 720\}$. The best results are in \textbf{bold}. $^*$ indicates results not reported in the original paper, and the results are reproduced.}
\resizebox{0.95\textwidth}{!}{
\begin{tabular}{ccc|cc|cc|cc|cc|cc}
\toprule
Model & \multicolumn{2}{c|}{\method} & \multicolumn{2}{c|}{CycleNet\tiny{/Linear}}  & \multicolumn{2}{c|}{CycleNet\tiny{/MLP}} & \multicolumn{2}{c|}{CATS} & \multicolumn{2}{c|}{PatchTST} & \multicolumn{2}{c}{iTransformer}  \\
\cmidrule(r){2-3} \cmidrule(r){4-5} \cmidrule(r){6-7} \cmidrule(r){8-9} \cmidrule(r){10-11} \cmidrule(r){12-13} 
Metric & MSE & MAE &  MSE & MAE &  MSE & MAE &  MSE & MAE &  MSE & MAE &  MSE & MAE   \\
\midrule
ETTh1&\textbf{0.426}&0.433&0.432&\textbf{0.427}&0.457&0.441&0.454&0.447&0.469&0.454&0.454&0.447\\
ETTh2&\textbf{0.378}&\textbf{0.402}&0.383&0.404&0.388&0.409&0.383&0.407&0.387&0.407&0.383&0.407\\
ETTm1&\textbf{0.378}&\textbf{0.394}&0.386&0.395&0.379&0.396&0.407&0.410&0.387&0.400&0.407&0.410\\
ETTm2&0.286&0.332&0.272&0.315&\textbf{0.266}&\textbf{0.314}&0.288&0.332&0.281&0.326&0.288&0.332\\
Exchange&\textbf{0.355}&\textbf{0.401}&0.383$^*$&0.414$^*$&0.378$^*$&0.410$^*$&0.374$^*$&0.409$^*$&0.367&0.404&0.360&0.403\\
Traffic&0.440&\textbf{0.278}&0.485&0.313&0.472&0.301&\textbf{0.428}&0.282&0.481&0.304&\textbf{0.428}&0.282\\
Weather&0.246&0.274&0.254&0.279&\textbf{0.243}&\textbf{0.271}&0.258&0.278&0.259&0.281&0.258&0.278\\
Electricity&0.176&0.266&0.170&0.260&\textbf{0.168}&\textbf{0.259}&0.178&0.270&0.205&0.290&0.178&0.270\\
Solar&0.224&\textbf{0.261}&0.235&0.270&\textbf{0.210}&\textbf{0.261}&0.233&0.262&0.270&0.307&0.233&0.262\\

\bottomrule
\end{tabular}
}
\label{tab:main96}
\end{center}
\end{table*}

\begin{table*}[htbp]
\centering
\caption{Forecasting results on the M4 dataset across different time granularities under SMAPE, MASE, and OWA metrics. Smaller values indicate better performance. \textbf{Bold} values denote the best average performance.}
\resizebox{0.85\textwidth}{!}{
\begin{tabular}{ccccccc}
\toprule
\multicolumn{2}{c}{Model} & \method & CATS & CycleNet/MLP & CycleNet/Linear & PatchTST \\
\midrule
\multirow{3}{*}{\rotatebox{90}{\small{Yearly}}}
& SMAPE & 13.364 $\pm$ 0.012 & 13.575 $\pm$ 0.007 & 13.534 $\pm$ 0.013 & 14.129 $\pm$ 0.002 & 13.529 $\pm$ 0.032 \\
& MASE & 3.003 $\pm$ 0.006 & 3.065 $\pm$ 0.008 & 3.056 $\pm$ 0.009 & 3.050 $\pm$ 0.001 & 3.042 $\pm$ 0.011 \\
& OWA & 0.786 $\pm$ 0.001 & 0.801 $\pm$ 0.002 & 0.798 $\pm$ 0.002 & 0.816 $\pm$ 0.000 & 0.796 $\pm$ 0.002 \\
\midrule
\multirow{3}{*}{\rotatebox{90}{\small{Quarterly}}}
& SMAPE & 10.202 $\pm$ 0.012 & 10.342 $\pm$ 0.013 & 10.310 $\pm$ 0.016 & 10.762 $\pm$ 0.001 & 10.843 $\pm$ 0.071 \\
& MASE & 1.201 $\pm$ 0.004 & 1.222 $\pm$ 0.008 & 1.215 $\pm$ 0.003 & 1.305 $\pm$ 0.002 & 1.292 $\pm$ 0.005 \\
& OWA & 0.902 $\pm$ 0.002 & 0.915 $\pm$ 0.001 & 0.911 $\pm$ 0.002 & 0.964 $\pm$ 0.001 & 0.963 $\pm$ 0.005 \\
\midrule
\multirow{3}{*}{\rotatebox{90}{\small{Monthly}}}
& SMAPE & 12.872 $\pm$ 0.019 & 13.750 $\pm$ 0.101 & 13.011 $\pm$ 0.018 & 13.346 $\pm$ 0.000 & 18.176 $\pm$ 0.482 \\
& MASE & 0.958 $\pm$ 0.003 & 1.071 $\pm$ 0.013 & 0.969 $\pm$ 0.002 & 1.015 $\pm$ 0.000 & 1.626 $\pm$ 0.110 \\
& OWA & 0.897 $\pm$ 0.002 & 0.980 $\pm$ 0.010 & 0.907 $\pm$ 0.001 & 0.940 $\pm$ 0.000 & 1.420 $\pm$ 0.209 \\
\midrule
\multirow{3}{*}{\rotatebox{90}{\small{Weekly}}}
& SMAPE & 9.791 $\pm$ 0.050 & 10.116 $\pm$ 0.243 & 12.025 $\pm$ 0.252 & 12.436 $\pm$ 0.267 & 9.821 $\pm$ 0.222 \\
& MASE & 3.166 $\pm$ 0.022 & 3.174 $\pm$ 0.102 & 3.887 $\pm$ 0.158 & 4.496 $\pm$ 0.108 & 3.157 $\pm$ 0.096 \\
& OWA & 1.079 $\pm$ 0.035 & 1.104 $\pm$ 0.029 & 1.356 $\pm$ 0.083 & 1.488 $\pm$ 0.049 & 1.134 $\pm$ 0.025 \\
\midrule
\multirow{3}{*}{\rotatebox{90}{\small{Daily}}}
& SMAPE & 3.063 $\pm$ 0.012 & 3.161 $\pm$ 0.119 & 3.113 $\pm$ 0.012 & 3.415 $\pm$ 0.000 & 3.322 $\pm$ 0.041 \\
& MASE & 3.286 $\pm$ 0.014 & 3.385 $\pm$ 0.122 & 3.354 $\pm$ 0.016 & 3.744 $\pm$ 0.000 & 3.532 $\pm$ 0.043 \\
& OWA & 1.004 $\pm$ 0.003 & 1.035 $\pm$ 0.038 & 1.023 $\pm$ 0.004 & 1.132 $\pm$ 0.000 & 1.0795 $\pm$ 0.014 \\
\midrule
\multirow{3}{*}{\rotatebox{90}{\small{Hourly}}}
& SMAPE & 18.802 $\pm$ 0.002 & 20.374 $\pm$ 0.125 & 18.335 $\pm$ 0.158 & 21.361 $\pm$ 0.000 & 21.734 $\pm$ 0.542 \\
& MASE & 2.769 $\pm$ 0.001 & 3.834 $\pm$ 0.137 & 2.939 $\pm$ 0.065 & 3.392 $\pm$ 0.000 & 4.065 $\pm$ 0.334 \\
& OWA & 1.090 $\pm$ 0.001 & 1.525 $\pm$ 0.045 & 1.168 $\pm$ 0.021 & 1.289 $\pm$ 0.000 & 1.439 $\pm$ 0.090 \\
\midrule
\multirow{3}{*}{\rotatebox{90}{\small{\textbf{Average}}}}
& SMAPE & \textbf{11.360} & 11.886 & 11.721 & 12.574 & 12.904 \\
& MASE & \textbf{2.422} & 2.625 & 2.570 & 2.833 & 2.785 \\
& OWA & \textbf{0.972} & 1.060 & 1.027 & 1.104 & 1.138 \\
\bottomrule
\end{tabular}
}
\label{tab:m4}
\end{table*}

\subsection{Results on M4 dataset}

In the previous sections, we have comprehensively evaluated \method on benchmark datasets of time series forecasting. However, these datasets span an excessively short time horizon (e.g., the weather dataset covers only a 1-year period, and ETT datasets cover only 2 years), rendering them inadequate for evaluating models specifically designed to capture seasonal patterns. To this end, we extend our evaluation to the M4 datasets from \url{https://www.kaggle.com/datasets/yogesh94/m4-forecasting-competition-dataset}, which contains time series data collected over years. The M4 dataset is divided into Yearly, Quarterly, Monthly, Weekly, Daily, and Hourly. The experimental results are reported in Table~\ref{tab:m4} using a significant test with five random seeds in \{2021,2022,2023,2024,2025\}. We have noticed that \method outperforms CycleNet on all six frequencies of the M4 dataset, with an overall improvement of 3.17\% (11.721 $\rightarrow$ 11.360) over CycleNet/MLP and 10.68\% (12.574 $\rightarrow$ 11.360) over CycleNet/Linear under the SMAPE metric. Moreover, \method achieves significant improvements on the yearly, quarterly, monthly, and hourly datasets.

\subsection{Ablation Study on More Dataset}

We conduct an additional ablation study on the Weather and ETTm2 dataset and show the results in the table~\ref{tab:abm}. We can draw three conclusions. First, incorporating attention bias consistently improves forecasting performance. Second, integrating periodic information also yields significant gains. Finally, using multiple periodic components is beneficial even in datasets without clear multi-periodicity, as fine-grained patterns (e.g., hourly) capture variations that coarser patterns (e.g., daily) miss.

\begin{table*}[ht]
\centering
\caption{More experiment results of different choices of periodic length. The look-back length $L$ is fixed as $336$ and the prediction horizons of $H \in \{96, 192, 336, 720\}$. Throughout this experiment, we set the number of attention heads to $12$ to ensure that the number of groups $g$ is a factor of $h$.}
\label{tab:ablation}
\resizebox{\textwidth}{!}{
\begin{tabular}{ccccccccccccccccc}
\toprule
\multicolumn{17}{c}{\textbf{Weather Dataset}} \\
\midrule
Case & \multicolumn{2}{c}{No Bias} & \multicolumn{2}{c}{Non-Periodic} & \multicolumn{6}{c}{Periodic} & \multicolumn{6}{c}{Both} \\
\cmidrule(l){2-3}\cmidrule(l){4-5}\cmidrule(l){6-11}\cmidrule(l){12-17}
$\mathcal{P}$ & \multicolumn{2}{c}{$\emptyset$} & \multicolumn{2}{c}{$\emptyset$} & \multicolumn{2}{c}{\{6\}} & \multicolumn{2}{c}{\{144\}} & \multicolumn{2}{c}{\{6,144\}} & \multicolumn{2}{c}{\{6\}} & \multicolumn{2}{c}{\{144\}} & \multicolumn{2}{c}{\{6,144\}} \\
\cmidrule(l){2-3}\cmidrule(l){4-5}\cmidrule(l){6-7}\cmidrule(l){8-9}\cmidrule(l){10-11}\cmidrule(l){12-13}\cmidrule(l){14-15}\cmidrule(l){16-17}
Metric & MSE & MAE & MSE & MAE & MSE & MAE & MSE & MAE & MSE & MAE & MSE & MAE & MSE & MAE & MSE & MAE \\
\midrule
96 & 0.151 & 0.199 & 0.151 & 0.200 & 0.151 & 0.200 & 0.150 & 0.202 & 0.149 & 0.201 & 0.152 & 0.201 & 0.151 & 0.202 & 0.150 & 0.200 \\
192 & 0.200 & 0.249 & 0.196 & 0.245 & 0.193 & 0.242 & 0.194 & 0.243 & 0.193 & 0.240 & 0.195 & 0.243 & 0.192 & 0.240 & 0.194 & 0.241 \\
336 & 0.252 & 0.287 & 0.250 & 0.284 & 0.246 & 0.282 & 0.245 & 0.284 & 0.244 & 0.281 & 0.243 & 0.281 & 0.243 & 0.282 & 0.243 & 0.282 \\
720 & 0.334 & 0.344 & 0.326 & 0.340 & 0.322 & 0.337 & 0.323 & 0.339 & 0.322 & 0.342 & 0.326 & 0.340 & 0.321 & 0.337 & 0.318 & 0.334 \\
\midrule
avg. & 0.234 & 0.270 & 0.230 & 0.267 & 0.228 & 0.265 & 0.228 & 0.267 & 0.227 & 0.266 & 0.229 & 0.266 & 0.227 & 0.265 & 0.226 & 0.264 \\
\midrule
\midrule
\multicolumn{17}{c}{\textbf{ETTm2 Dataset}} \\
\midrule
Case & \multicolumn{2}{c}{No Bias} & \multicolumn{2}{c}{Non-Periodic} & \multicolumn{6}{c}{Periodic} & \multicolumn{6}{c}{Both} \\
\cmidrule(l){2-3}\cmidrule(l){4-5}\cmidrule(l){6-11}\cmidrule(l){12-17}
$\mathcal{P}$ & \multicolumn{2}{c}{$\emptyset$} & \multicolumn{2}{c}{$\emptyset$} & \multicolumn{2}{c}{\{4\}} & \multicolumn{2}{c}{\{96\}} & \multicolumn{2}{c}{\{4,96\}} & \multicolumn{2}{c}{\{4\}} & \multicolumn{2}{c}{\{96\}} & \multicolumn{2}{c}{\{4,96\}} \\
\cmidrule(l){2-3}\cmidrule(l){4-5}\cmidrule(l){6-7}\cmidrule(l){8-9}\cmidrule(l){10-11}\cmidrule(l){12-13}\cmidrule(l){14-15}\cmidrule(l){16-17} 
Metric & MSE & MAE & MSE & MAE & MSE & MAE & MSE & MAE & MSE & MAE & MSE & MAE & MSE & MAE & MSE & MAE \\
\midrule
96 & 0.174 & 0.265 & 0.172 & 0.262 & 0.167 & 0.257 & 0.168 & 0.258 & 0.168 & 0.257 & 0.167 & 0.257 & 0.172 & 0.261 & 0.171 & 0.260 \\
192 & 0.229 & 0.301 & 0.226 & 0.298 & 0.223 & 0.294 & 0.227 & 0.296 & 0.224 & 0.294 & 0.224 & 0.294 & 0.224 & 0.294 & 0.224 & 0.295 \\
336 & 0.288 & 0.337 & 0.285 & 0.334 & 0.279 & 0.333 & 0.282 & 0.333 & 0.279 & 0.333 & 0.278 & 0.332 & 0.278 & 0.331 & 0.282 & 0.332 \\
720 & 0.378 & 0.391 & 0.373 & 0.390 & 0.370 & 0.388 & 0.374 & 0.390 & 0.370 & 0.388 & 0.370 & 0.389 & 0.370 & 0.390 & 0.371 & 0.391 \\
\midrule
avg. & 0.267 & 0.324 & 0.264 & 0.321 & 0.260 & 0.318 & 0.263 & 0.319 & 0.260 & 0.318 & 0.260 & 0.318 & 0.261 & 0.319 & 0.262 & 0.320 \\
\bottomrule
\end{tabular}
}
\label{tab:abm}
\end{table*}

\subsection{Ablation Study of Patching Stride}

To elaborate on the performance of various patching strides in \method, we conduct comprehensive experiments of different patching lengths $P$ and stride $S$ illustrated in Table ~\ref{tab:stridemore}. The results consistently demonstrate that reducing the stride $S$ enhances prediction accuracy across multiple datasets and metrics. For instance, in the ETTh1 dataset, a smaller stride significantly improves performance, with the average MSE decreasing from 0.426 at $P=8, S=4$ to 0.419 at $P=1, S=1$, with 1.64\% decrease. Similarly, the trend is also evident for the Weather dataset, where the smallest stride configuration ($P=4, S=1$) achieves the lowest MSE.
Overall, shorter strides allow \method to process a greater number of overlapping patches, enabling finer-grained feature extraction and better long-range dependency modeling. 
\begin{table}[htbp]
\centering

\resizebox{\linewidth}{!}{
\begin{tabular}{c|c|cc|cc|cc|cc|cc|cc|cc|cc|cc}
\toprule
\multicolumn{2}{c}{P/S} & \multicolumn{2}{c|}{16/8} & \multicolumn{2}{c|}{8/8} & 
\multicolumn{2}{c|}{8/4} & \multicolumn{2}{c|}{4/4} & \multicolumn{2}{c|}{4/2} & \multicolumn{2}{c|}{2/2} & \multicolumn{2}{c|}{4/1} & \multicolumn{2}{c|}{2/1} & \multicolumn{2}{c}{1/1}  \\

\cmidrule(r){3-4} \cmidrule(r){5-6} \cmidrule(r){7-8} \cmidrule(r){9-10} \cmidrule(r){11-12} \cmidrule(r){13-14} 
\cmidrule(r){15-16} \cmidrule(r){17-18}  \cmidrule(r){19-20} 
\multicolumn{2}{c}{Metric} & MSE & MAE & MSE & MAE & MSE & MAE & MSE & MAE & MSE & MAE & MSE & MAE & MSE & MAE & MSE & MAE &MSE & MAE \\
\midrule
\multirow{5}{*}{\rotatebox{90}{ETTh1}} 
&96&0.382&0.401&0.382&0.400&0.380&0.397&0.381&0.398&0.379&0.397&0.378&0.397&0.379&0.398&0.378&0.396&0.380&0.400\\
&192&0.416&0.421&0.416&0.419&0.417&0.418&0.418&0.418&0.416&0.418&0.414&0.417&0.422&0.423&0.416&0.418&0.418&0.424\\
&336&0.435&0.434&0.437&0.435&0.437&0.433&0.434&0.429&0.441&0.438&0.432&0.430&0.459&0.455&0.435&0.434&0.431&0.436\\
&720&0.459&0.474&0.466&0.478&0.466&0.478&0.458&0.470&0.454&0.466&0.455&0.467&0.446&0.460&0.463&0.472&0.431&0.436\\
\cmidrule(l){2-20}
& avg. & 0.423&0.433&0.425&0.433&0.426&0.432&0.423&0.429&0.423&0.430&0.420&0.428&0.427&0.434&0.423&0.430&\textbf{0.419}&\textbf{0.430}\\
\midrule
\multirow{5}{*}{\rotatebox{90}{ETTh2}} 
&96&0.287&0.344&0.289&0.345&0.290&0.346&0.288&0.344&0.287&0.345&0.287&0.343&0.291&0.347&0.291&0.347&0.290&0.347\\
&192&0.354&0.387&0.355&0.387&0.357&0.389&0.354&0.388&0.355&0.389&0.362&0.396&0.359&0.393&0.359&0.393&0.354&0.389\\
&336&0.380&0.410&0.380&0.410&0.379&0.410&0.379&0.412&0.380&0.413&0.378&0.410&0.396&0.425&0.396&0.425&0.379&0.412\\
&720&0.397&0.433&0.399&0.436&0.398&0.434&0.394&0.432&0.399&0.437&0.400&0.438&0.392&0.430&0.392&0.430&0.397&0.435\\
\cmidrule(l){2-20}
& avg. & 0.355&0.394&0.356&0.395&0.356&0.395&\textbf{0.354}&\textbf{0.394}&0.355&0.396&0.357&0.397&0.360&0.399&0.360&0.399&0.355&0.396 \\
\midrule
\multirow{5}{*}{\rotatebox{90}{ETTm1}} 
&96&0.293&0.344&0.301&0.345&0.286&0.339&0.293&0.345&0.281&0.337&0.291&0.342&0.293&0.348&0.295&0.347&0.299&0.351\\
&192&0.332&0.370&0.335&0.371&0.329&0.367&0.327&0.365&0.319&0.363&0.327&0.365&0.322&0.365&0.334&0.372&0.332&0.372\\
&336&0.362&0.387&0.361&0.388&0.357&0.387&0.356&0.387&0.353&0.385&0.355&0.387&0.353&0.386&0.354&0.386&0.356&0.389\\
&720&0.416&0.418&0.419&0.421&0.405&0.418&0.404&0.416&0.403&0.416&0.401&0.416&0.406&0.418&0.399&0.415&0.407&0.420\\
\cmidrule(l){2-20}
& avg. &0.351&0.380&0.354&0.381&0.344&0.378&0.345&0.378&\textbf{0.339}&\textbf{0.375}&0.344&0.378&0.344&0.379&0.346&0.380&0.349&0.383\\
\midrule
\multirow{5}{*}{\rotatebox{90}{Weather}} 
&96&0.150&0.198&0.152&0.203&0.150&0.199&0.152&0.202&0.152&0.203&0.151&0.202&0.148&0.200&0.200&0.195&0.157&0.207\\
&192&0.195&0.242&0.197&0.244&0.193&0.240&0.195&0.244&0.193&0.243&0.198&0.245&0.194&0.241&0.242&0.245&0.197&0.245\\
&336&0.248&0.282&0.247&0.281&0.246&0.281&0.246&0.283&0.246&0.282&0.249&0.283&0.243&0.282&0.282&0.319&0.249&0.285\\
&720&0.324&0.334&0.321&0.333&0.323&0.337&0.320&0.334&0.320&0.334&0.319&0.333&0.320&0.337&0.319&0.333&0.321&0.336\\
\cmidrule(l){2-20}
& avg. &0.229&0.264&0.321&0.333&0.228&0.264&0.228&0.266&0.228&0.266&0.229&0.266&\textbf{0.226}&0.265&0.227&\textbf{0.264}&0.231&0.268 \\
\bottomrule

\end{tabular}}
\caption{Additional experiment results of the impact of various patching strides. The look-back length $L$ is fixed as $336$ and the prediction horizons of $H \in \{96, 192, 336, 720\}$. In the table, P indicates the patching length, while S indicates the stride.}
\label{tab:stridemore}

\end{table}





\section{Additional Robustness Study}

\subsection{Robustness Towards Varied Look-Back Window}

\begin{figure*}[htbp]
    \centering
    \includegraphics[width=0.96\linewidth]{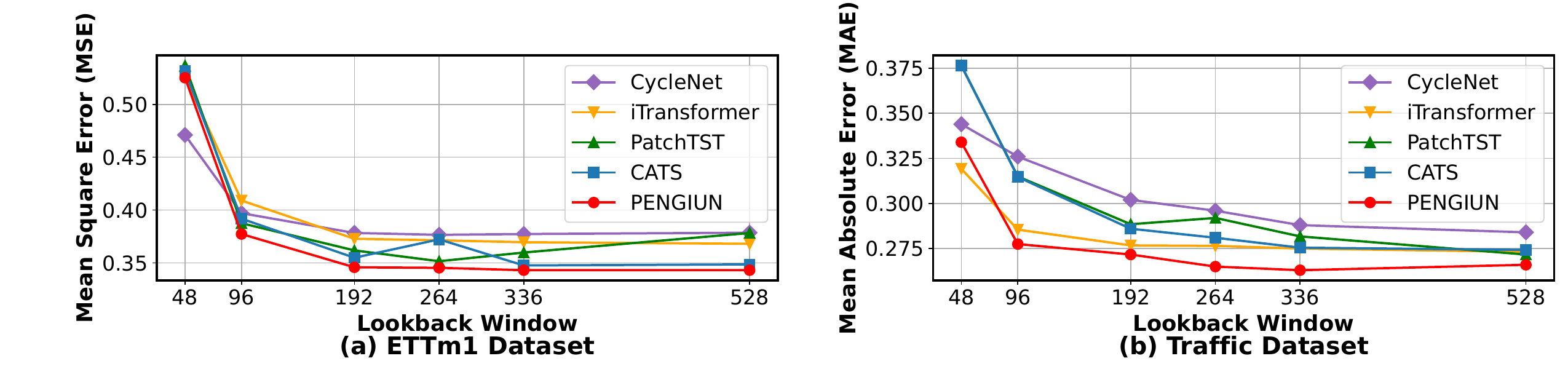}
    \caption{The experimental results of various models with diverse lookback windows $L$ on the ETTm1 and Traffic dataset. The experiments are averaged over the predicted horizon of $H \in \{96,192,336,720\}$. }
    \label{fig:lookback}
\end{figure*}

To assess \method's robustness, we conducted experiments on the ETTm1 and Traffic datasets, varying the look-back window length, denoted as $L$, across a range of values: $[48,96,192,264,336,528]$. As shown in the experimental results in Figure~\ref{fig:lookback}, \method consistently outperforms baseline models across these different window lengths when the input length is larger than $96$. We hypothesize that an input length of $48$ is insignificant for capturing temporal patterns. \method's ability to effectively handle varying temporal scales and long-range dependencies in time series data. 

\begin{table*}[t]
\centering
\caption{Robustness analysis with five random seeds compared to baselines. The look-back length $L$ is fixed
as $336$ and the results are averaged over prediction horizons of $H \in \{96, 192, 336, 720\}$.}
\resizebox{0.85\textwidth}{!}{
\begin{tabular}{cccccccc}
\toprule
\multicolumn{2}{c}{Model} & \multicolumn{2}{c}{\method} & \multicolumn{2}{c}{CycleNet/Linear} & \multicolumn{2}{c}{PatchTST} \\
\cmidrule(l){3-4}\cmidrule(l){5-6}\cmidrule(l){7-8}
\multicolumn{2}{c}{Metric} & MSE & MAE & MSE & MAE & MSE & MAE \\
\midrule
\multirow{4}{*}{\rotatebox{90}{Traffic}} 
& 96 & 0.357 $\pm$ 0.000 & 0.247 $\pm$ 0.000 & 0.399 $\pm$ 0.000 & 0.276 $\pm$ 0.000 & 0.378 $\pm$ 0.002 & 0.270 $\pm$ 0.002 \\
& 192 & 0.376 $\pm$ 0.000 & 0.253 $\pm$ 0.000 & 0.413 $\pm$ 0.000 & 0.282 $\pm$ 0.000 & 0.394 $\pm$ 0.002 & 0.274 $\pm$ 0.003 \\
& 336 & 0.390 $\pm$ 0.000 & 0.262 $\pm$ 0.000 & 0.426 $\pm$ 0.000 & 0.288 $\pm$ 0.000 & 0.406 $\pm$ 0.003 & 0.283 $\pm$ 0.004 \\
& 720 & 0.428 $\pm$ 0.000 & 0.286 $\pm$ 0.000 & 0.453 $\pm$ 0.000 & 0.305 $\pm$ 0.000 & 0.443 $\pm$ 0.004 & 0.298 $\pm$ 0.004 \\
\midrule
\multirow{4}{*}{\rotatebox{90}{Solar}} 
& 96 & 0.180 $\pm$ 0.002 & 0.245 $\pm$ 0.002 & 0.202 $\pm$ 0.001 & 0.286 $\pm$ 0.001 & 0.195 $\pm$ 0.004 & 0.252 $\pm$ 0.003 \\
& 192 & 0.202 $\pm$ 0.005 & 0.257 $\pm$ 0.006 & 0.225 $\pm$ 0.001 & 0.297 $\pm$ 0.000 & 0.201 $\pm$ 0.003 & 0.263 $\pm$ 0.002 \\
& 336 & 0.203 $\pm$ 0.006 & 0.252 $\pm$ 0.003 & 0.227 $\pm$ 0.001 & 0.261 $\pm$ 0.000 & 0.208 $\pm$ 0.005 & 0.274 $\pm$ 0.004 \\
& 720 & 0.213 $\pm$ 0.005 & 0.257 $\pm$ 0.003 & 0.237 $\pm$ 0.001 & 0.264 $\pm$ 0.000 & 0.211 $\pm$ 0.004 & 0.268 $\pm$ 0.003 \\
\midrule
\multirow{4}{*}{\rotatebox{90}{ETTm1}}
& 96 & 0.290 $\pm$ 0.001 & 0.341 $\pm$ 0.000 & 0.301 $\pm$ 0.000 & 0.344 $\pm$ 0.000 & 0.289 $\pm$ 0.001 & 0.345 $\pm$ 0.001 \\
& 192 & 0.319 $\pm$ 0.000 & 0.361 $\pm$ 0.000 & 0.335 $\pm$ 0.000 & 0.364 $\pm$ 0.000 & 0.339 $\pm$ 0.002 & 0.378 $\pm$ 0.002 \\
& 336 & 0.356 $\pm$ 0.001 & 0.387 $\pm$ 0.000 & 0.369 $\pm$ 0.000 & 0.384 $\pm$ 0.000 & 0.373 $\pm$ 0.002 & 0.400 $\pm$ 0.001 \\
& 720 & 0.409 $\pm$ 0.000 & 0.418 $\pm$ 0.001 & 0.424 $\pm$ 0.000 & 0.414 $\pm$ 0.000 & 0.434 $\pm$ 0.004 & 0.438 $\pm$ 0.003 \\
\bottomrule
\end{tabular}
}
\label{tab:mainrc}
\end{table*}

\subsection{Robustness Towards Random Seeds}

In addition, we conducted robustness tests used in the main comparison experiments. To validate the stability of our approach against random parameter initialization, we evaluated \method with five different random seeds \{2021,2022,2023,2024,2025\}. Table \ref{tab:mainrc} reports the standard deviation of MSE and MAE for different methods across these seeds. Results show that \method achieves statistically significant improvements over strong baselines on these datasets. 

\section{Limitation and Future Work} 

\method demonstrates its effectiveness in LTSF with the proposed periodic causal attention. However, there are several potential limitations of \method that warrant discussion here. 
Firstly, \method might not be suitable for datasets with varied periodic lengths over time, such as electrocardiogram (ECG) data, since \method employs a predefined set of periodic lengths. Secondly, \method, as well as other Transformer-based methods, typically requires more GPU memory and computation cost compared to linear models, because the attention matrix is square in relation to the input length. In future work, we plan to develop flash attention to conserve GPU memory and explore linear attention approaches to reduce computational costs.

\end{document}